\documentclass[journal]{IEEEtran}
\usepackage{amsmath,amsfonts}
\usepackage{algorithmic}
\usepackage{algorithm}
\usepackage{array}
\usepackage[caption=false,font=normalsize,labelfont=sf,textfont=sf]{subfig}
\usepackage{textcomp}
\usepackage{stfloats}
\usepackage{url}
\usepackage{verbatim}
\usepackage{graphicx}
\usepackage{cite}
\usepackage{booktabs}
\usepackage{multirow}
\usepackage{bbm}
\usepackage{diagbox}
\usepackage[colorlinks,
            linkcolor=blue,
            anchorcolor=blue,
            citecolor=blue]{hyperref}
\usepackage[section]{placeins}

\begin{document}

\title{Bridging Data Islands: Geographic Heterogeneity-Aware Federated Learning for Collaborative Remote Sensing Semantic Segmentation}

\author{Jieyi Tan, Yansheng Li,~\IEEEmembership{Senior Member,~IEEE,} Sergey A. Bartalev, Shinkarenko Stanislav, Bo Dang, Yongjun Zhang,~\IEEEmembership{Member,~IEEE,} Liangqi Yuan, Wei Chen

\thanks{This research was supported by the National Natural Science Foundation of China under Grant 42371321 and Grant 42030102. (Corresponding author: Yansheng Li.)}
\thanks{Jieyi Tan, Yansheng Li, Bo Dang and Yongjun Zhang are with the School of Remote Sensing and Information Engineering, Wuhan University, Wuhan 430079, China (email: tanjieyi@whu.edu.cn; yansheng.li@whu.edu.cn; bodang@whu.edu.cn; zhangyj@whu.edu.cn)}
\thanks{Sergey A. Bartalev and Shinkarenko Stanislav are with the Space Research Institute, Russian Academy of Sciences, Moscow 119421, Russia (email: bartalev@d902.iki.rssi.ru; wise\_snake@bk.ru)}
\thanks{Liangqi Yuan is with the College of Engineering, Purdue University, West Lafayette, IN 47907, USA (email: liangqiy@purdue.edu)}
\thanks{Wei Chen is with the Institute of Surface-Earth System Science, School of Earth System Science, Tianjin University, Tianjin 300072, China (email: chenwei19@tju.edu.cn)}
}

\markboth{Journal of \LaTeX\ Class Files,~Vol.~14, No.~8, August~2021}%
{Shell \MakeLowercase{\textit{et al.}}: A Sample Article Using IEEEtran.cls for IEEE Journals}



\maketitle

\begin{abstract}
Remote sensing semantic segmentation is an essential technology in earth observation missions. Due to concerns over geographic information security, data privacy, storage bottleneck and industry competition, high-quality annotated remote sensing images are often isolated and distributed across various institutions. The issue of remote sensing data islands poses challenges for fully utilizing isolated datasets to train a global model. Federated learning (FL), a privacy-preserving distributed collaborative learning technology, offers a potential solution to leverage large-scale isolated remote sensing data. Typically, remote sensing images from different institutions exhibit significant geographic heterogeneity, characterized by coupled class-distribution heterogeneity and object-appearance heterogeneity. However, existing FL methods lack consideration of them, leading to a decline in the performance of the global model when FL is directly applied to remote sensing semantic segmentation. To cope with the aforementioned geographic heterogeneity, we propose a novel \underline{Geo}graphic heterogeneity-aware \underline{Fed}erated learning (GeoFed) framework to bridge data islands in remote sensing semantic segmentation. Our framework consists of three modules, including the Global Insight Enhancement (GIE) module, the Essential Feature Mining (EFM) module and the Local-Global Balance (LoGo) module, which collaboratively optimize the model in an end-to-end manner. The geographic heterogeneity is alleviated through the synergies of them. Through the GIE module, class distribution heterogeneity is alleviated by introducing a prior global class distribution vector. Additionally, we design an EFM module to alleviate object appearance heterogeneity by constructing essential features, which are further optimized by intra-institution and inter-institution contrastive constraints. Furthermore, the LoGo module enables the model to possess both global generalization capability and local adaptation. Extensive experiments on three public datasets (i.e., FedFBP, FedCASID, FedInria) demonstrate that our GeoFed framework consistently outperforms the current state-of-the-art methods. Specifically, It achieves improvements in global performance by 2.50\%, 2.17\%, and 1.03\%, and in average local performance by 1.62\%, 2.40\%, and 0.55\%, respectively.
\end{abstract}


\newpage
\section{Introduction}\label{intro}
\IEEEPARstart{L}{and} use and land cover (LULC) mapping products can serve as foundational information and support decision making processes in sustainable development goals (SDGs)~\cite{ISPRS_LULC,RSE2024deepULU}. Typically, creating or updating LULC mapping products involves collecting data on individual regions by local authorities and mapping by experts. Remote sensing stands out for its extensive spatial coverage, frequent updates, and cost-effectiveness, addressing the challenges associated with intensive data collection from large-scale areas. Researchers have been devoted to producing and updating LULC products through remote sensing interpretation~\cite{ISPRS_LULC}.

Nowadays, remote sensing semantic segmentation has been a prevalent solution in automatic LULC mapping tasks~\cite{RSE_global_landuse}. With the rapid development of deep learning, this data-driven paradigm has demonstrated excellent performance in remote sensing semantic segmentation. As is well known, deep learning-based remote sensing semantic segmentation highly relies on large-scale, high-quality, and fine-grained remote sensing datasets. However, remote sensing images with annotations tend to be distributed across institutions. They are collected, stored, managed and utilized privately because remote sensing data sharing is still hindered by geospatial information security, storage bottleneck, and industry competition~\cite{GRSM_Aisecurity,GRSM_FL}. Consequently, the dilemma of remote sensing data islands emerges.

\begin{figure*}[t]    
  \centering           
  \subfloat[]  
  {
      \label{fig:motivation_a}\includegraphics[width=0.45\textwidth]{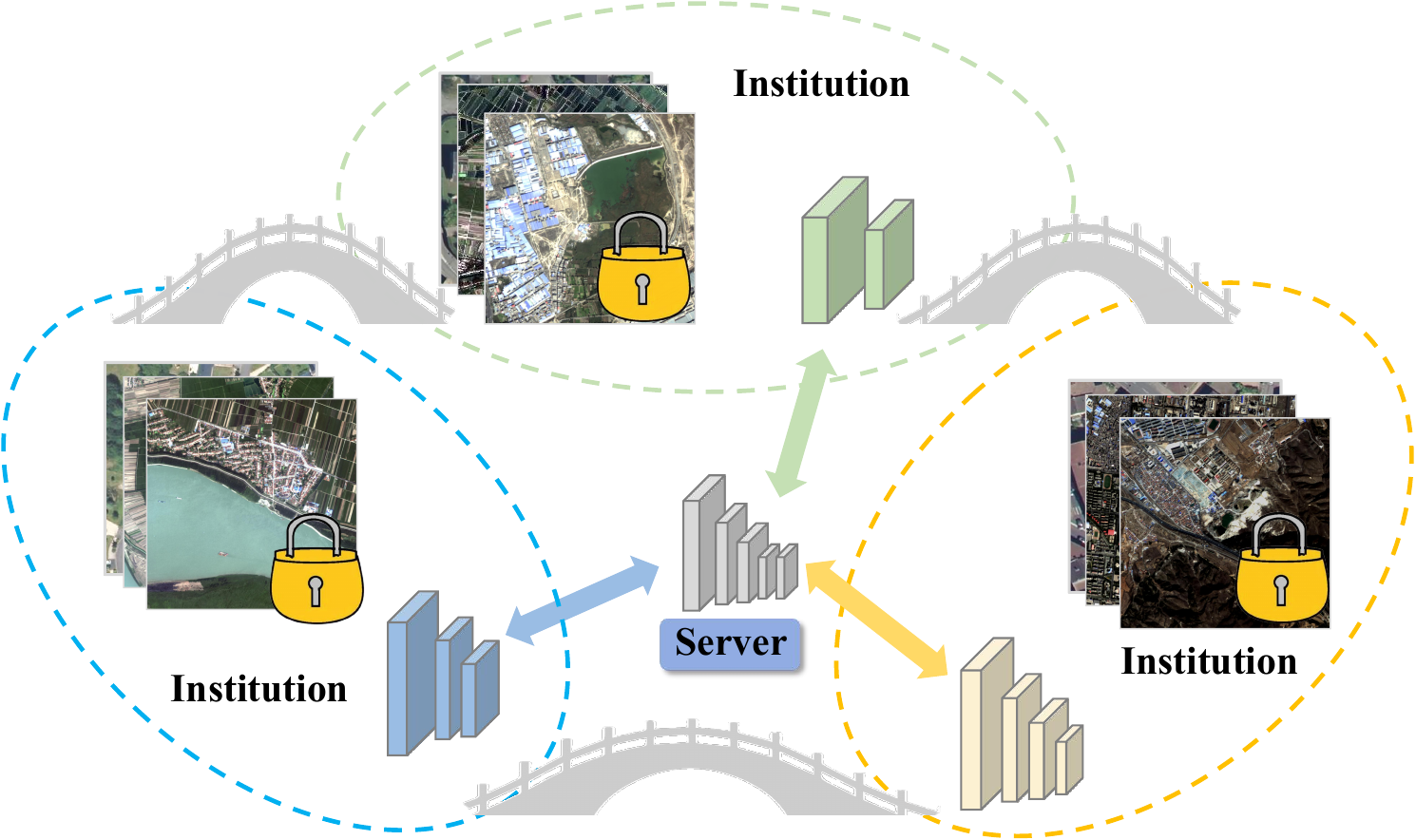}
  }
\hspace{0cm}
  \subfloat[]
  {
      \label{fig:motivation_b}\includegraphics[width=0.5\textwidth]{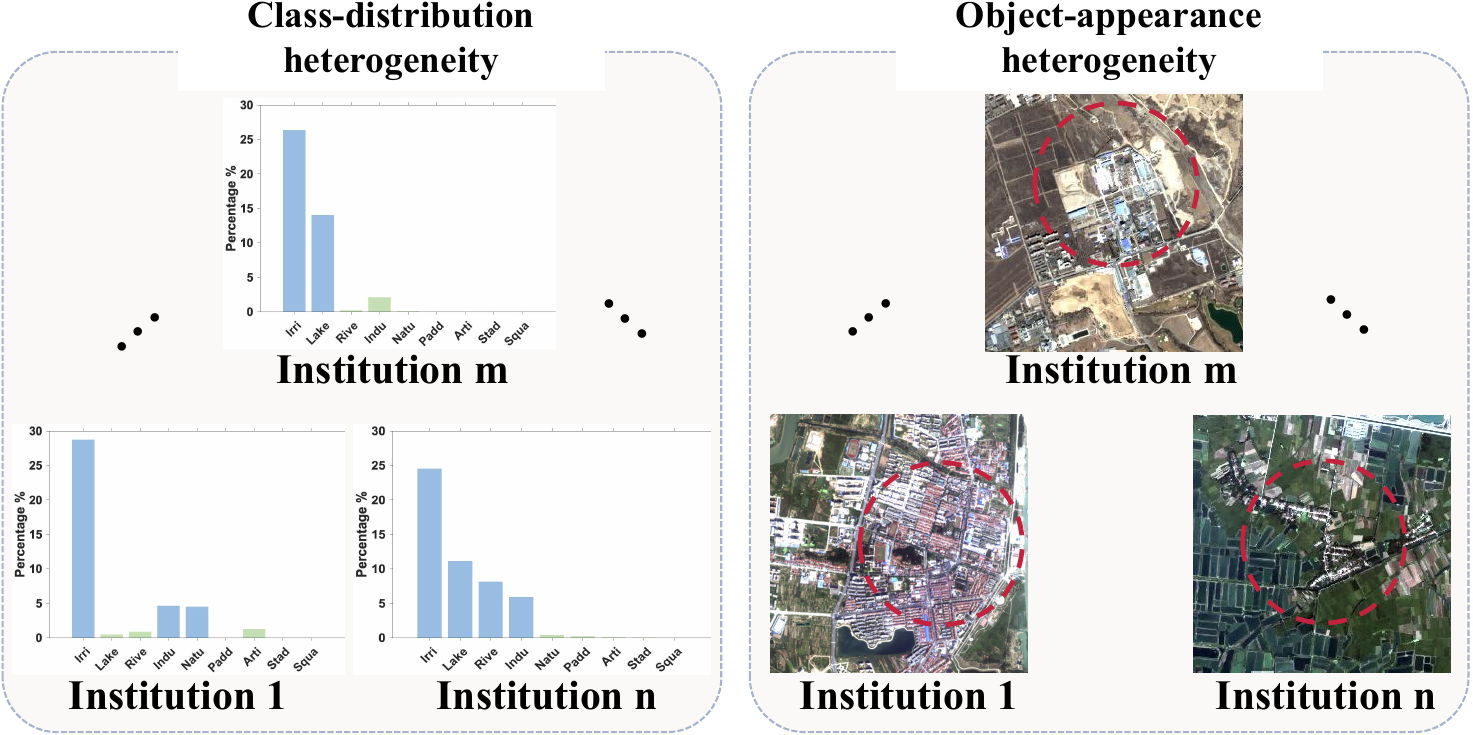}
  }
  \caption{\textbf{(a) Illustration of bridging remote sensing data islands.} Institutions only transmit model parameters without disclosing their private data, therefore achieving privacy-preserving collaborative learning. \textbf{(b) Traditional FL applied in remote sensing semantic segmentation encounters geographic heterogeneity.} The coexistence of class distribution heterogeneity and object appearance heterogeneity limits the model performance.}    
  \label{fig:motivation}           
\end{figure*}

In the context of data islands, local data on islands face the challenge of scarcity. In contrast, there is a growing demand for large-scale remote sensing semantic segmentation for LULC mapping tasks on a global~\cite{RSE_global_scale}, national~\cite{IJDE_national_scale}, and regional~\cite{RSE_region_scale} scale. Recently, trustworthy artificial intelligence in remote sensing interpretation (TRSI) has gained increasing attention from researchers~\cite{ISPRS_trustworthy}. Bridging remote sensing data islands is a crucial challenge in achieving TRSI (Fig.~\ref{fig:motivation}\subref{fig:motivation_a}). Despite its significance, there is still a lack of a comprehensive exploration of the privacy-preserving remote sensing semantic segmentation. As well known, federated learning (FL) is a decentralized paradigm that promotes collaborative learning among clients~\cite{AISTATS_FedAvg,shao2024selective} In FL, each institution only needs to train locally and exchange models without sharing its raw data. Hence, FL is promising to address the challenges of remote sensing data islands but is scarcely explored in remote sensing semantic segmentation. Fig.~\ref{fig:cen&FL} illustrates the differences between the traditional remote sensing semantic segmentation and the novel FL paradigms.

As the second law of geography says, \textit{geographic variables exhibit uncontrolled variance}~\cite{second_law}. The formation of geographic landscapes is a complex interplay of various factors, such as climate, geology, hydrology, biodiversity, and human activities. These factors shape the unique characteristics of different regions~\cite{ISPRS_landuse_hetero,RSE_hetero_landscape}. Firstly, remote sensing data from various institutions are typically collected from different regions, naturally resulting in geographic heterogeneity among institutions. Therefore, when optimizing locally, each institution tends to focus on its own local optimum, which is inconsistent with the optimum direction of the global model constructed through FL, making it difficult for the FL model to converge to the global optimum. Secondly, the global model often struggles to adapt to local institution's characteristics, leading to a discrepancy between global and local performance. Recently, many studies~\cite{CVPR_re_hetero, MLSys_FedProx, CVPR_FedSeg,ECCV_flat} are dedicated to addressing the issue of heterogeneity in FL. However, the existing studies on the effective application of FL in remote sensing semantic segmentation remain insufficient and calls for further investigation and systematic discussions. Therefore, the advantages of FL in remote sensing semantic segmentation are not well demonstrated. Most of the works do not fully consider the data characteristics in remote sensing. From a more comprehensive perspective, both class-distribution and object-appearance heterogeneity exist in the remote sensing data among the institutions. In Fig.~\ref{fig:motivation}\subref{fig:motivation_b}, we present the challenges brought about by geographic heterogeneity. A detailed explanation of the two types of heterogeneity reflected in remote sensing semantic segmentation is as follows:

\begin{figure*}[htbp]    
  \centering           
  \subfloat[]  
  {
      \label{fig:centralized}\includegraphics[width=0.45\textwidth]{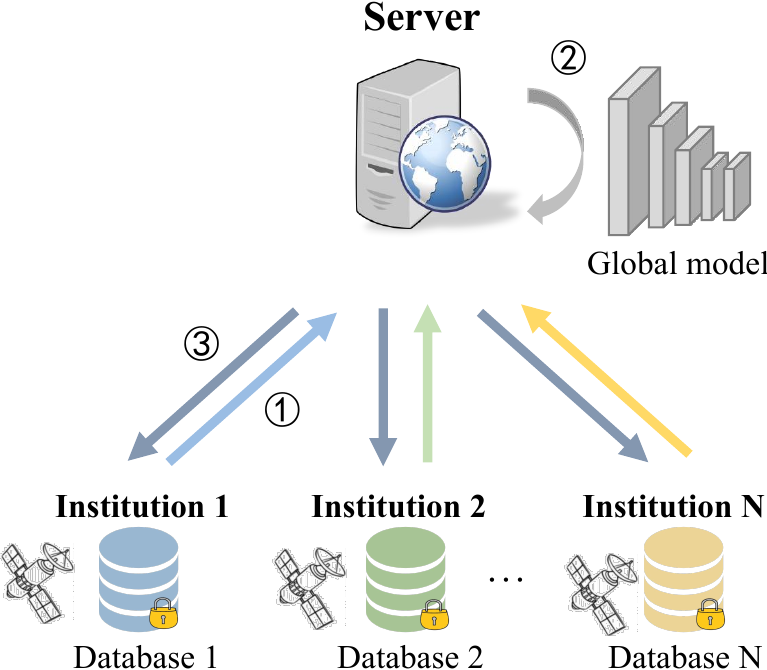}
  }
  \hspace{1cm}
  \subfloat[]
  {
      \label{fig:FL}\includegraphics[width=0.45\textwidth]{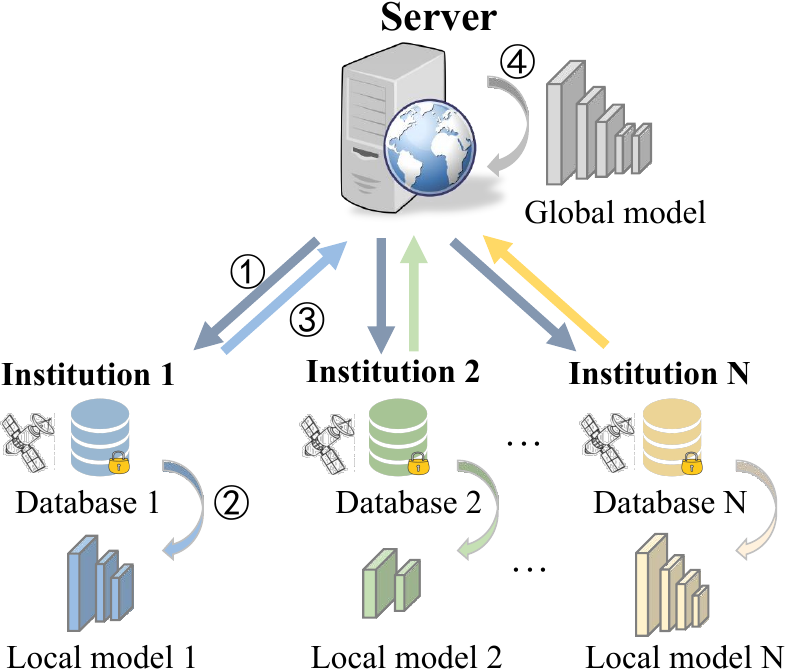}
  }
  \caption{\textbf{(a)} \textbf{Illustration of the traditional remote sensing semantic segmentation training paradigm.} \textcircled{1} All collaborating institutions upload their private remote sensing data to the server for centralized storage. \textcircled{2} The server conducts centralized training. \textcircled{3} The server distributes the trained models to the respective institutions. \textbf{(b)} \textbf{Illustration of the FL paradigm for remote sensing semantic segmentation.} \textcircled{1} Distribution of the global model. \textcircled{2} Update of local models. \textcircled{3} Upload of local models. \textcircled{4} global models aggregation. }    
  \label{fig:cen&FL}           
\end{figure*}

\textbf{(1) Class-Distribution Heterogeneity}: As is shown in the left part of Fig.~\ref{fig:motivation}\subref{fig:motivation_b}, each local dataset grapples with a class imbalance problem. Moreover, local class distributions vary and do not align with the global class distribution. Additionally, class imbalance in remote sensing images typically presents as a long-tailed distribution~\cite{TPAMI_LT_survey}. In extreme cases, data in some institutions may lack certain classes. This situation intensifies the class imbalance problem, especially for minority class samples~\cite{ICCV_ls_semi,ESSD_SinoLC1}. For example, due to a combination of various regional factors, the number of common geographic features significantly outnumbers uncommon ones. Institutions in tropical regions may have an abundance of forests but little to no snow. Conversely, polar regions may have more snow. From a global perspective, snow then becomes a minority category. The class distributions of each local institution and the global ones exhibit considerable heterogeneity.


\textbf{(2) Object-Appearance Heterogeneity}: In the right part of Fig.~\ref{fig:motivation}\subref{fig:motivation_b}, different regions have different characteristics for the same land cover category, which becomes more evident in coarse-grained classification systems~\cite{ISPRS_FBP,RSE_LCZ_mapping,CVPR_dynamic}. The objects exhibit heterogeneous appearances. For example, in rainy areas, there is a tendency to construct houses with pointed roofs for better drainage compared to dry areas. As a result, the overall feature representations of the buildings differ across different institutions. The global model needs to learn the essential features (e.g., regular shape of buildings) while also maintaining robustness to diverse institution-specific features (e.g., flat or pointed roofs)~\cite{ACM_heter}. 

Overall, the complexity increases further as geographic heterogeneity is often characterized by the coexistence of class distribution heterogeneity and object appearance heterogeneity mentioned above. Furthermore, inconsistent class taxonomies often arise from variations in geographic distribution. Due to variations in imaging conditions caused by the atmosphere and the sensitivity of remote sensing data to both temporal and spatial changes, FL for remote sensing semantic segmentation encounters additional uncertainties, making the problem more challenging. From a positive perspective, the geographic heterogeneity inherent in each institution's remote sensing data naturally shapes the mapping zone pattern. It simplifies complex surface features into several types of homogeneous regions, significantly compressing the information representation of surface features. Zone mapping facilitates individual institutions in exploring local geographic knowledge while collaborating with others to learn generalized knowledge. Addressing inconsistency in optimization goals arising from geographic heterogeneity becomes particularly crucial in this process.

Considering the aforementioned challenges, we propose a \underline{Geo}graphic heterogeneity-aware \underline{Fed}erated Learning (GeoFed) framework to formulate a novel paradigm for privacy-preserving collaborative remote sensing semantic segmentation. To address the geographic heterogeneity challenges posed in FL for remote sensing semantic segmentation, the GeoFed framework is designed with three components. (i) Class-distribution heterogeneity is alleviated through the Global Insight Enhancement (GIE) module. (ii) Essential Feature Mining (EFM) strategy to alleviate object-appearance heterogeneity by constructing essential features. (iii) Local-Global Balance Policy (LoGo) enables the model to possess both global generalization capability and local data personalization. 
The main contributions of this paper can be summarized as follows:

\begin{enumerate}
\item We propose a novel privacy-preserving remote sensing semantic segmentation framework by geographic heterogeneity-aware federated learning. To the best of our knowledge, our work is the first comprehensive exploration of the challenges posed by geographic heterogeneity in FL for remote sensing semantic segmentation.

\item This paper comprehensively considers remote sensing semantic segmentation-oriented geographic heterogeneity in privacy-preserving learning. We propose the GIE module to alleviate class distribution heterogeneity. An EFM module is proposed to alleviate object appearance heterogeneity. Lastly, the LoGo module enables the models to possess both global generalization capability and local data personalization.  

\item Extensive experiments on three public datasets (i.e., FedFBP, FedCASID, FedInria) with typical geographic heterogeneity show our GeoFed framework outperforms state-of-the-art methods on three datasets simultaneously.
\end{enumerate}

The remainder of this paper is organized as follows. Section~\ref{related} introduces the related work in heterogeneous FL and privacy-preserving remote sensing interpretation. Section~\ref{method} describes the details of our GeoFed framework. The experimental results are analyzed in Section~\ref{exper}, and Section~\ref{conclusion} concludes this paper.

\section{Related work}\label{related}

This section briefly reviews the most relevant works regarding the aforementioned aspects, including heterogeneity in FL and some other privacy-preserving techniques for remote sensing interpretation.

\subsection{Heterogeneous FL}
The pioneering FedAvg~\cite{AISTATS_FedAvg} trained a global model by aggregating local model parameters. In each communication round, all institutions receive the aggregated model parameters and conduct the \textit{Local Update} procedure in parallel. Subsequently, the server aggregates the optimized model parameters from institutions into a single model, which is used in the next communication round.~\cite{ICLR_convergence} demonstrated that the naive FedAvg algorithm converges under non-independent and identically distributed (non-IID) data scenarios.
However, in scenarios with heterogeneous data, its performance will significantly decrease~\cite{ACM_heter,ECCV_flat,IJCAI_CReFF,ICDE_experi,yuan2024communication}. 

Recently, many studies have proposed methods to deal with this challenge mainly from three perspectives: local updates, global aggregation, and model structure design. FedSeg~\cite{CVPR_FedSeg} proposed a framework to address the class heterogeneity in federated semantic segmentation, but it focuses more on addressing the foreground-background inconsistency problem, which is not a key issue in remote sensing semantic segmentation. FISS~\cite{CVPR_FISS} studied how to achieve incremental learning in federated semantic segmentation, with a particular focus on addressing the catastrophic forgetting issue caused by heterogeneous data. Some other methods adopted strategies from domain generalization~\cite{CVPR_FedDG,CVPR_DaFKD,CVPR_Style_DG,IROS_auto_drive}. 
GBME~\cite{ICCV_GBME_FedLT}  designed a proxy as the
class prior for re-balancing algorithms without requiring additional private information, but it can not be applied to the more challenging semantic segmentation task.
Some studies utilized prototype learning to solve the problem~\cite{IOT_FedMargin,CVPR_rethinking_prototype}. Some studies investigated the architecture of federated models, and observed that heterogeneity can be alleviated by Transformers~\cite{CVPR_re_archi,ICCV_FRAug}. Moreover, Some other works focused on the aggregation procedures. Elastic aggregation~\cite{CVPR_Elastic} reduced the magnitudes of updates to the more sensitive parameters to prevent the server model from drifting to any one institution distribution, and conversely boosted updates to the less sensitive parameters to better explore different institution distributions. FedKTL~\cite{Zhang_2024_CVPR} leveraged the knowledge stored in public pre-trained generators to deal with the knowledge sharing difficulities in heterogeneous FL.

FL for remote sensing semantic segmentation aims to transfer knowledge between different clients to learn models with superior performance collaboratively. However, the aforementioned works did not fully consider the data characteristics (i.e., the combined influence of dual geographic heterogeneity patterns) in remote sensing semantic segmentation. Class-distribution and object-appearance heterogeneity cause knowledge transfer barriers.

\subsection{Privacy-Preserving Remote Sensing Interpretation}
Remote sensing interpretation requires a large amount of sensitive remote sensing data. Privacy protection in remote sensing interpretation has gradually brought attention along with applying deep learning techniques~\cite{GRSM_Aisecurity,chen2024free}.

In literature, some pioneering studies built FL frameworks on Unmanned Aerial Vehicle (UAV) swarms, which
enables different devices to collaboratively monitor without sharing raw data~\cite{ACM_aerial,IoT_FL_sky,RS_DP}. SACDF~\cite{TGRS_fedhyper} proposed an improved FL for the hyperspectral classification task, using the multidimensional spatial details of hyperspectral images to perceive decision boundaries. Some studies explored the application of FL in remote sensing image classification~\cite{JSTARS_FL_image_classification,TGRS_Radar}, but these methods consider classification tasks rather than the more challenging semantic segmentation tasks. Some recent studies validated FL on remote sensing semantic segmentation datasets, but they neglect the geographic heterogeneity of remote sensing images~\cite{TGRS_FedPM,IGARSS_proto_IAIL,ICO_FSS}. FedPM~\cite{TGRS_FedPM} proposed an FL method based on prototype matching for object extraction in remote sensing, but it did not consider the problem of multi-class segmentation and ignores the class-distribution heterogeneity in remote sensing semantic segmentation. FSDCL~\cite{GRSL_FS} focused on-orbit updating and promote on-orbit model performance under data-scarcity conditions. MPCL~\cite{TGRS_MPL} constructed a regularized strategy based framework which fosters multiple parties cooperatively process multiple remote sensing tasks. \cite{Arxiv_FedDiff} proposed a diffusion model-driven FL for multi-modal which is vertical FL. Different from these existing privacy-preserving remote sensing interpretation works, our GeoFed framework demonstrates a strong awareness of geographic heterogeneity, enabling effective improvement in collaborative remote sensing semantic segmentation performance.

\section{Methodology}\label{method}
In this section, we introduce the proposed GeoFed
framework and its main components in sequence, starting with the problem definition~(Section~\ref{subsec:problem}) and an overview of our GeoFed framework~(Section~\ref{subsec:overview}).

\subsection{Problem Definition}
\label{subsec:problem}
Suppose that there are a total of $n$ institutions participating in FL, and the $i$-th institution owns its private remote sensing semantic segmentation dataset $\mathcal{D}^{i}$: $\{x_{m}^{i}, y_{m}^{i}\}_{m=1}^{|\mathcal{D}^{i}|}$. Each pixel $x_{m}^{i}$ has a corresponding label of category $y_{m}^{i}$. Specifically, $\mathcal D^1,\cdots,\mathcal D^n$ are collected and sampled from distinctive geographic distributions. The entire dataset can be expressed as $\mathcal{D}=\{\mathcal{D}^{1}, \mathcal{D}^{2}, \cdots, \mathcal{D}^{n}\}$. Assume there are a total of $C$ categories in the dataset, and let ${f}_{c}^{i}$ denote the label frequency of class $c$ in the $i$-th institution. 

In the traditional paradigm, the whole dataset $\mathcal{D}$ is used to conduct centralized training. In the FL paradigm, we aim for each institution to train a semantic segmentation model $w$ with good generalization and local personalization without transferring the local dataset $\mathcal{D}^{i}$. The objective function can be formulated as follows:

\begin{equation}
\underset{w}{\arg \min}\mathcal{L}(w)=\sum_{i=1}^n\frac{|\mathcal{D}^i|}{|\mathcal{D}|}\mathcal{L}_i(w),
\label{eq:FL_obejective}
\end{equation}
where $|\mathcal{D}|$ denotes the number of samples in $\mathcal{D}$, and $\mathcal{L}_i(w)$ denotes the empirical loss of the $i$-th institution. It is formulated as follows:

\begin{equation}
\mathcal{L}_i(w)=\mathbb{E}_{(\mathbf{x},\mathbf{y})\in \mathcal{D}^i}\ell_i [(\mathbf{x},\mathbf{y});w],
\label{eq:empirical}
\end{equation}
where $\ell_i$ is the local loss of the model $w$ in the $i$-th institution for the remote sensing image $\mathbf{x}$ and its corresponding mask $\mathbf{y}$ in the local dataset $\mathcal{D}^i$. 

The data distribution of the $i$-th and $j$-th institution are denoted as ${\mathbb{P}}_{i}(x,y)$ and ${\mathbb{P}}_{i}(x,y)$, respectively, which can be rewritten as ${\mathbb{P}}_{i}(x|y){\mathbb{P}}_{i}(y)$ and ${\mathbb{P}}_{j}(x|y){\mathbb{P}}_{j}(y)$. For class distribution heterogeneity, ${\mathbb{P}}_{i}(y)\neq {\mathbb{P}}_{j}(y), (i\neq j)$, in extreme cases, some institutions may lack categories. For object appearance heterogeneity, ${\mathbb{P}}_{i}(x|y)\neq {\mathbb{P}}_{j}(x|y), (i\neq j)$, this means that objects of the same category present different appearance in different institutions.

\subsection{Overview of Our GeoFed Framework}
\label{subsec:overview}
\begin{figure*}
    \centering
    \includegraphics[width=1\linewidth]{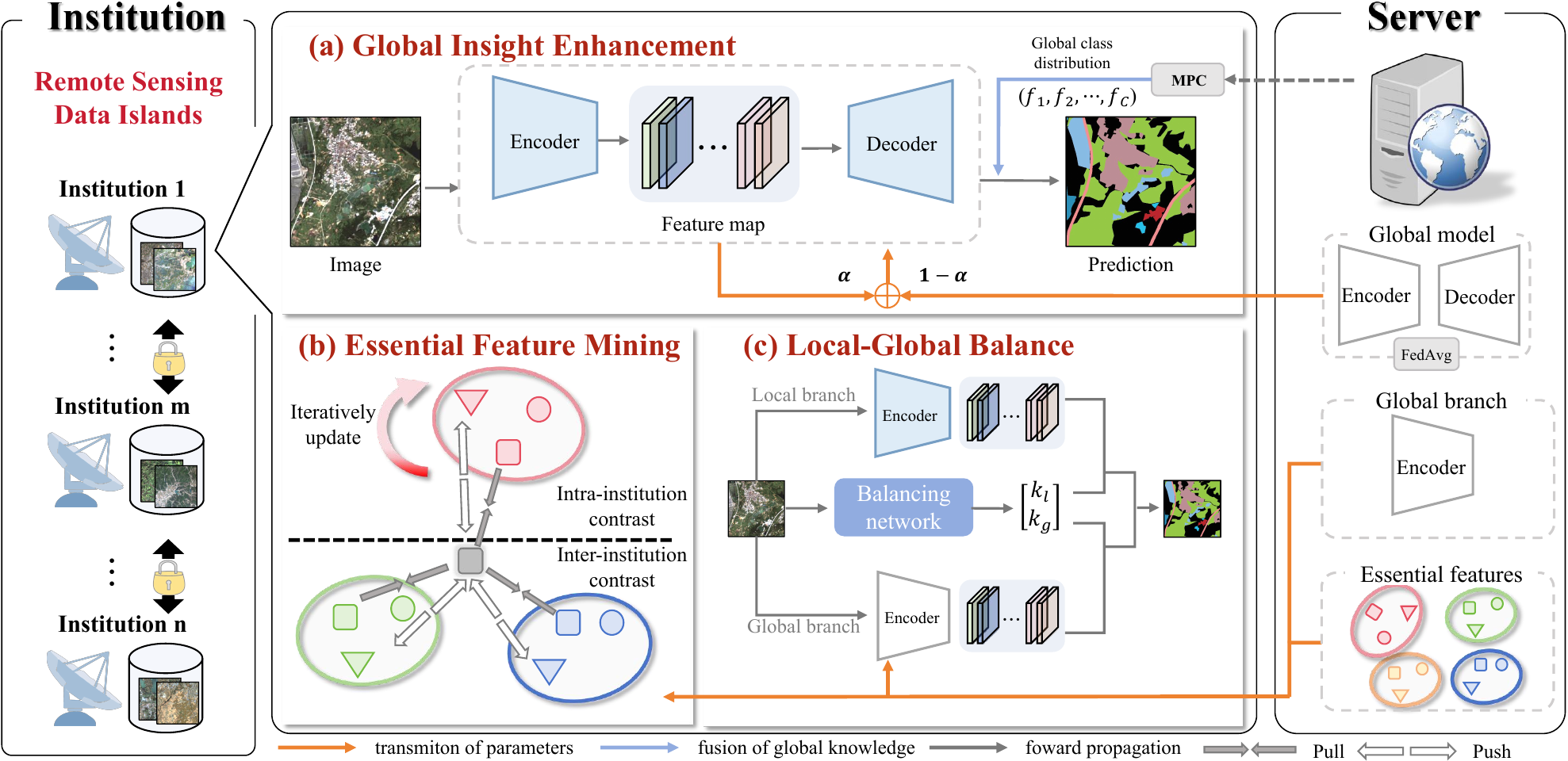}
    \caption{\textbf{An overview of our proposed GeoFed framework.} It mainly contains three components. Firstly, (a) class-distribution heterogeneity is alleviated through the utilization of the GIE module. GIE expands the feature diversity of local models under the global class distribution and injects the global information. Next, (b) an EFM module containing intra \& inter contrastive loss is applied to alleviate object-appearance heterogeneity. Lastly, (c) a LoGo module with three branches is applied to achieve the balance between local characteristics and global generalization.}
    \label{fig:overview}
\end{figure*}

\begin{figure}
    \centering
    \includegraphics[width=1\linewidth]{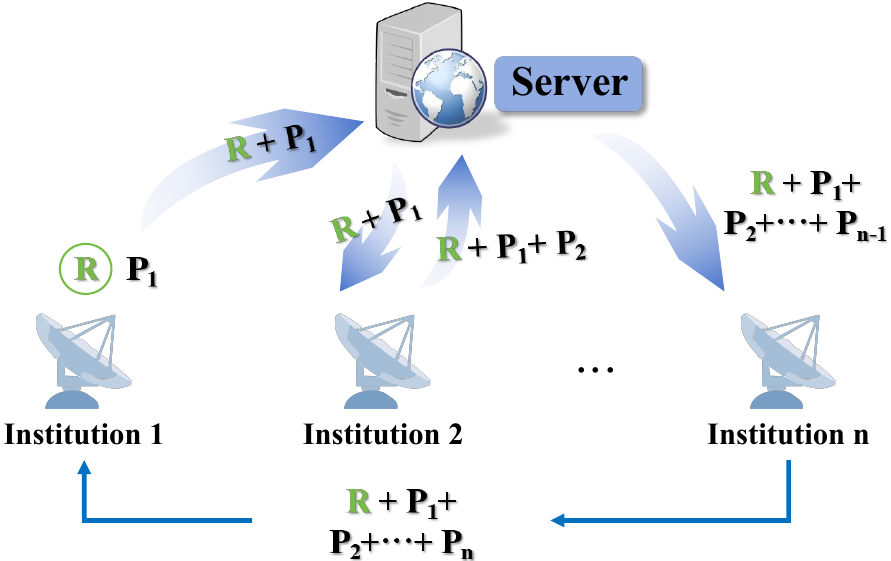}
    \caption{\textbf{The MPC process schematic diagram.} $R$ represents a relatively large private random number generated by the first institution, and $P_i$ denotes the local private class distribution vectors of the i-th institution. Institutions sequentially add their local class distribution vectors in this process and pass the results to the next institution. The last institution directly transmits the result to the first institution. After subtracting R, the first institution broadcasts the global class distribution to all participants.}
    \label{fig:MPC}
\end{figure}

An overview of our GeoFed framework is presented in Fig~\ref{fig:overview}. To alleviate the issue of class-distribution heterogeneity, we propose the GIE module~(Section~\ref{subsec:GIE}), which enables institutions to learn more diverse features and aligns them with the global class distribution, thus mitigating class imbalance. To address the problem of object-appearance heterogeneity, we propose the EFM module~(Section~\ref{subsec:EFM}), which utilizes contrastive loss to facilitate the learning of more compact and discriminative features by institution models and encourages the discovery of common essential features across institutions. The adaptability of a single global model to geographic heterogeneity data from various institutions is insufficient. We propose a LoGo module to enable the model to possess global generalization capability and local data adaptation~(Section~\ref{subsec:LoGo}). These modules will be elaborated in detail in the following sections.

The total loss function of the i-th institution for the proposed GeoFed framework is written as follows:

\begin{equation}
 \mathcal{L}_i=\mathcal{L}_{CE}+\lambda_{1}\mathcal{L}_{inter}+\lambda_{2}\mathcal{L}_{intra},
\label{eq:loss}
\end{equation}
where $\mathcal{L}_{CE}$ is from the the GIE module. The details of $\mathcal{L}_{inter}$ and $\mathcal{L}_{intra}$ will be elaborated in the EFM module. The LoGo module plays a role in framework integration and is not involved in the composition of the total loss function. Moreover, $\lambda_{1}$ and $\lambda_{2}$ are hyper-parameters used to control the inter-contrastive and intra-contrastive loss weights. 

In the following sections, the architecture of the GIE module is first
explained in detail. We then describe the EFM module and LoGo,
focusing on how they handle the geographic heterogeneity issues.

\begin{figure*}[t]
    \centering
    \includegraphics[width=1\linewidth]{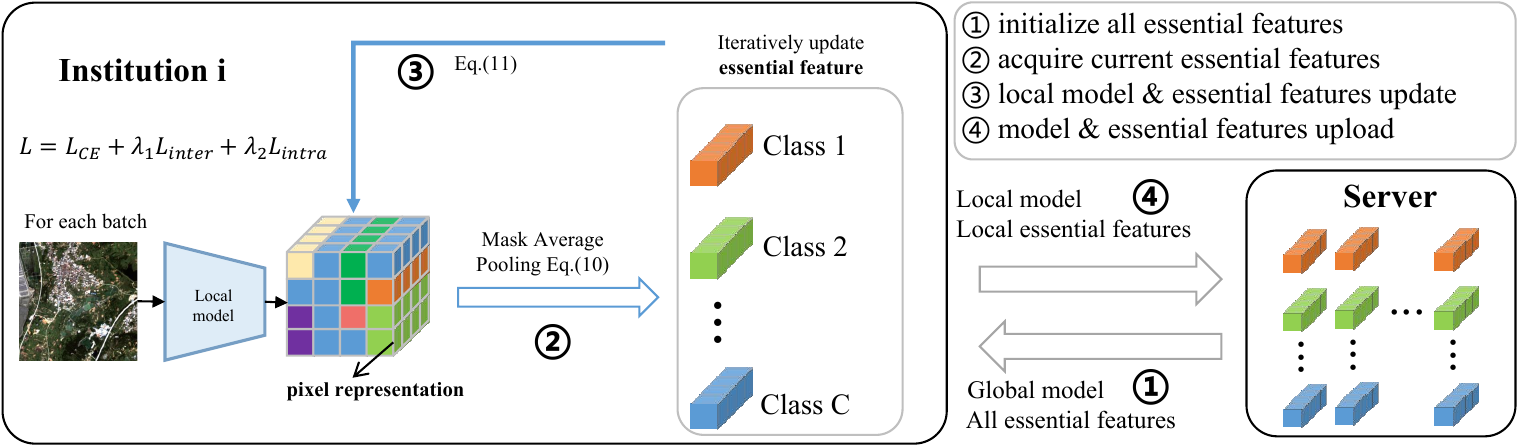}
    \caption{\textbf{Illustration of Essential Feature Mining.}}
    \label{fig:EFM}
\end{figure*}
\label{subsec:EFM}

\subsection{Global Insight Enhancement}
\label{subsec:GIE}

Classes with fewer labels often lack robust feature representations due to the insufficient sample size. On the institution's side, we propose diversity enhancement for global tail categories. Firstly, we calculate the class frequency vector of each institution using the following formula:
\begin{equation}
f_{c}=\frac{\sum_{i=1}^{N}\sum_{j=1}^{H\times W}y^{i,j,c}}{N\times H\times W},
\label{eq:frequency}
\end{equation}
where $N$, $W$, $H$ denote the number of batches, width, and height of the image, respectively. The global class frequency $f_c$ can be obtained by calculating the local frequency vectors of each institution. 

Each pixel corresponds to a feature vector $q_i$ and a true class label $y_{i}^{c}$. We assign smaller Gaussian perturbations to head categories and larger Gaussian perturbations to tail categories. The formula is as follows: 
\begin{equation}
q_{i}^{'} = q_i + w_{{c}}|\delta(\sigma)|,
\label{eq:pert}
\end{equation}
where $\delta(\sigma)$ follows a Gaussian distribution with mean 0 and standard deviation $\sigma$. We denote $f_{c'} =\frac{1}{f_{c}+\epsilon}$, $\epsilon$ is for ensure not divided by zero. $w_{{c}}$ is the perturbation scale, and is normalized by~\eqref{eq:scale}:
\begin{equation}
S_{{c}} = \frac{\exp{f_{c'}}}{\sum_{c=1}^{C}\exp{f_{c'}}}.
\label{eq:scale}
\end{equation}

The loss is computed using the conventional Cross-Entropy loss on the feature map after applying Gaussian perturbations, as shown in~\eqref{eq:CEloss},
\begin{equation}
\mathcal{L}_{CE} = -\sum_{c=1}^{C}y_{c}^{i}\log{q^i}.
\label{eq:CEloss}
\end{equation}

It is worth noting that privacy leakage of local class distribution is not a concern here, as the global frequency vector can be easily obtained without leaking local frequency vector information using Secure Multi-party Computation (MPC) technology~\cite{ACM_MPC}. The schematic diagram is shown in Fig~\ref{fig:MPC}.

The most extreme manifestation of the long-tail distribution is the phenomenon of Broken Tail, where some institutions experience category absence, thereby exacerbating the issue of class imbalance. For a given institution $m$, we regard categories with a total pixel percentage less than $\tau$ as Broken Tail. The number of remaining categories is signified as $C_{residue}$. 

 Drawing inspiration from EWF~\cite{CVPR_EWF}, we design a simple yet efficient Tail Regeneration (TR) strategy analogous to the anti-forgetting mechanism, which effectively integrates the knowledge of a more knowledgeable global model and an institution model, mitigating the performance degradation of missing classes caused by class distribution heterogeneity. It further introduces global information to the local update process of our GeoFed framework. 
 
 For every aggregated global model ($\omega_{global}$), the model becomes more experienced via the aggregation process and generally possesses a better recognition ability for each category. The updated model ($\omega_{update}$) post-local training at the institution's end tends to overfit the remaining categories. We introduce the parameter $\alpha$ to preserve the original global model's insights. The TR strategy can be expressed as

\begin{equation}
\alpha_{TR}=\alpha \odot \omega_{update}+(1-\alpha)\odot \omega_{global},
\label{eq:TR_fusion}
\end{equation}
where $\odot$ is Hadamard product and $\alpha$ is defined as follows:
\begin{equation}
\alpha=\sqrt{\dfrac{C_{\textit{residue}}}{C+C_{\textit{residue}}}}.
\label{eq:percent}
\end{equation}

\subsection{Essential Feature Mining}

The object-appearance heterogeneity leads to inconsistencies in the optimization objectives during the training processes across various institutions. The introduction of the EFM is aimed at fostering FL for remote sensing semantic segmentation to mine the essential features of land cover categories.
In Fig~\ref{fig:EFM}, we provide a more detailed illustration of the workflow of it.

The essential feature of the category $c$ in the $i$-th institution is obtained using masked average pooling~\cite{ICCV_MAP} based on the representation of pixel $j$, denoted as $P_{i}(j)$. 
\begin{equation}
p_{c,i}^t=\frac{\sum_{j}P_{i}(j)\mathbbm{1}\left[y_{i}(j)=c\right]}{\sum_{j}\mathbbm{1}\left[y_{i}(j)=c\right]}, 
\label{eq:MAP}
\end{equation}
where $\mathbbm{1}$ is the indicator function, and $t$ is the number of communication round. The essential features $p_{c,i}^t$ are updated online and initialized as a vector following the Gaussian distribution of $\cal{N}$(0,1). The formula for online updating is as follows:
\begin{equation}
p_{c,i}^t= \gamma p_{c,i}^{t}  + (1-\gamma)p_{c,i}^{t-1}
\label{eq:online}
\end{equation}

Each institution uploads its updated model to the server along with the essential features extracted locally. Since these essential features are just a few vectors and do not contain the raw remote sensing data, there is very low risk of privacy leakage. To better explore essential features and promote the model to learn common features, we design intra-contrastive loss and inter-contrastive loss based on Info-NCE loss~\cite{ICML_InfoNCE}. The EFM module is designed to fully utilize existing supervision signals in remote sensing objects from various regions, encouraging institutions' intra-class compactness and inter-class separability, leading to better consistent representations.

For intra-contrastive loss, the objective is to promote the proximity of feature representations belonging to the same categories within each institution, fostering a cohesive grouping. Simultaneously, the loss seeks to maximize the separation between feature representations corresponding to distinct categories, emphasizing distinctiveness. The intra-contrastive loss can be written as:

\begin{equation}
{\mathcal{L}}_{intra}=\sum_{}-l o g{\frac{\exp(p_{j}p_{c+}/\tau)}{\exp(p_{j}p_{c+}/\tau)+\sum_{p_{c-}}\exp(p_{j}p_{c-}/\tau)}},
\label{eq:contrast_intra}
\end{equation}
where $p_{j}^{i}$ denotes the normalized representation vector of a sample pixel in the $i$-th institution. $p_{c+}$ denotes the essential feature of the same class, and $p_{c-}$ is of the different classes. $\tau$ is the temperature parameter.

For inter-contrastive loss, we aim to make the pixel feature representations of a certain category in one institution close to the essential features of the same category in other institutions while being far from the essential features of different categories in other institutions. This enables each institution to optimize towards the global essential features and alleviates the issue of object-appearance heterogeneity. The inter-contrastive loss is formulated as:
\begin{equation}
{\mathcal{L}}_{inter}={\frac{1}{n}}\sum_{i=1}^{n}-l o g{\frac{\exp(p_{j}^{i}p_{c+}/\tau)}{\exp(p_{j}^{i}p_{c+}/\tau)+\sum_{p_{c-}}\exp(p_{j}^{i}p_{c-}/\tau)}}, 
\label{eq:contrast_inter}
\end{equation}
where $p_{j}^{i}$ denotes the normalized representation vector of a sample pixel in the $i$-th institution. $p_{c+}$ denotes the essential feature of the same class, and $p_{c-}$ is of the different classes. $\tau$ is the temperature parameter.

\begin{algorithm}[t]
\caption{Our GeoFed Framework}
\label{alg:GeoFed}
\begin{algorithmic}[1] 
\REQUIRE Total number of institutions $n$, total communication rounds $T$, local learning rate $\eta_l$, balancing branch learning rate $\eta_b$, local training epoch $E$.
\ENSURE Global remote sensing semantic segmentation model $F_g({\omega_g})$.
\STATE \textbf{Initialization:} Initialize the global model $F_g({\omega_g})$ and the balancing network $B^i({\omega_b})$.
\vspace{0.5em}

\STATE \textbf{Local Update:}
\FOR{$e = 0$ to $E - 1$}
    \STATE Download global model parameters ${\omega}_g$.
    \STATE Calculate local class frequency. \hfill {\eqref{eq:frequency}}
    \STATE Compute global $f_c \gets \text{MPC()}$.
    \STATE Compute cross-entropy loss $\mathcal{L}_{CE}$ with perturbations. \hfill {\eqref{eq:pert}}
    \STATE Update essential features online. \hfill {\eqref{eq:MAP}, \eqref{eq:online}}
    \STATE Compute contrastive losses $\mathcal{L}_{inter}$ and $\mathcal{L}_{intra}$. \hfill {\eqref{eq:contrast_intra}, \eqref{eq:contrast_inter}}
    \STATE Combine losses: $\mathcal{L} = \mathcal{L}_{CE} + \lambda_{1}\mathcal{L}_{inter} + \lambda_{2}\mathcal{L}_{intra}$. \hfill {\eqref{eq:loss}}
    \STATE Perform tail regeneration. \hfill {\eqref{eq:TR_fusion}}
\ENDFOR
\vspace{0.5em}

\STATE \textbf{Server Execution:}
\FOR{$t = 0$ to $T - 1$}
    \STATE Distribute global model $\omega_g^{(t)}$ to each institution.
    \FOR{each institution $i = 1$ to $n$ \textbf{in parallel}}
        \STATE Update local model ${\omega_{i}^{(t)}} \gets \text{LocalUpdate}(i, \omega_g^{(t)})$.
        \STATE Update essential features $\gets \text{LocalUpdate}(i, \omega_g^{(t)})$.
        \STATE Perform Local-Global Balance module. \hfill {\eqref{eq:end}}
    \ENDFOR
    \STATE Aggregate global model: ${\omega^{(t+1)}_{g}} = \sum_{i=1}^n \frac{|\mathcal{D}^i|}{|\mathcal{D}|} {\omega_{i}^{(t)}}$.
\ENDFOR
\end{algorithmic}
\end{algorithm}

\subsection{Local-Global Balance}
\label{subsec:LoGo}
The adaptability of a single global model to geographic heterogeneity data from various institutions is insufficient. In response, we propose a Local-Global Balance (LoGo) module to enable the model to possess global generalization capability and local data adaptation. Following local updates to the model and the global model, we seek a balanced model that fits the local data distribution. Leveraging a ResNet18 network to train local data, we map images through the model to a binary classification problem. Thus, the output logits represent the weights of the updated and global models. Through continuous optimization of the network, we obtain the optimal hybrid model of the local and global models.

LoGO consists of three branches, each serving a distinct purpose in the process of local training updates across institutions. The local branch is responsible for mining local knowledge, while the global branch maintains model generalization. The balancing network adopts a lightweight ResNet-18 to achieve local and global knowledge balancing. The local branch network is represented as $F_{l}(\omega_{l}) = F_{l}(\omega_{l}^{en}) \circ F_{l}(\omega_{l}^{de})$, the global branch network as $F_{g}(\omega_{g}) = F_{g}(\omega_{g}^{en}) \circ F_{g}(\omega_{g}^{de})$, and the balancing network as $B_{\theta}(\omega_{b})$. The input $\mathbf{x}\in \mathbb{R}^{N\times C\times H\times W}$, where $N$ denotes the batch size, and $C\times H\times W$ represent the channels, height, and width of the image, respectively. We elaborate the details of three branched as follows:

\textbf{(1) Local Branch.}
The remote sensing image is fed into the local branch. This branch adopts GIE and EFM modules elaborated above. Through the local branch, the local representation $\mathcal{R}_l\in \mathbb{R}^{N\times C_l\times H_l\times W_l}$ is denoted as: $\mathcal{R}_l = F_{l}(\mathbf{x};\omega_{l}^{en})$.

\textbf{(2) Global Branch.}
The global branch network is updated by aggregation of local branches in each round of aggregation and distributed to all institutions. The aggregation update formula is as follows: 
\begin{equation}
{\omega^{(t+1)}_{g}} = \sum_{i=1}^n\frac{|\mathcal{D}^i|}{|\mathcal{D}|}{\omega_{l_i}^{(t)}}.
\label{eq:aggregation}
\end{equation}

The encoder of the global branch is frozen. The image is fed into the global branch. The representation of the global branch $\mathcal{R}_g\in \mathbb{R}^{N\times C_l\times H_l\times W_l}$ is denoted as $\mathcal{R}_g = F_{g}(\mathbf{x};\omega_{g}^{en})$.

\textbf{(3) Balancing Network.}
To achieve a balance between global knowledge and local knowledge. We use a balancing network to produce the weight between them, resulting in a two-dimensional balancing coefficient. The representation of knowledge balance mixing is obtained by inputting the balanced representation into the decoder, simultaneously optimizing end-to-end losses for the local and balancing branches.

\begin{equation}
    \begin{aligned} 
        &\left[K_l; K_g\right] = B_{\theta}(\mathbf{x}; \omega_{b}), \\
       & \mathcal{R}_b = K_l *\mathcal{R}_l + K_g *\mathcal{R}_g
    \end{aligned}
    \label{eq:co}
\end{equation}

\textbf{(4) End-to-End Optimization.} The entire framework is integrated. The models are optimized in an end-to-end as follows:
\begin{equation}
  \left\{
    \begin{gathered}
        \omega_l^{(t)} \gets \omega_l^{(t-1)} - \eta_l \nabla \mathcal{L}_i \\
    \omega_b^{(t)} \gets \omega_b^{(t-1)} - \eta_b\nabla \mathcal{L}_b,
    \end{gathered}
  \right.
  \label{eq:end}
\end{equation}
where $\eta_l$ is the learning rate of the local branch, and $\eta_b$ is the learning rate of the balancing branch.

To provide a clearer overview of our GeoFed framework, the overall pipeline of this distributed system is shown in Algorithm~\ref{alg:GeoFed}.

\section{Experiments and discussions}\label{exper}
In this section, we first introduce the dataset, evaluation metrics and the experiment settings. Afterward, we compare our GeoFed framework with other state-of-the-art frameworks on three re-organized public datasets. Finally, we demonstrate the effectiveness of the geographic heterogeneity-aware design with qualitative results. In the following subsections, the experimental settings, comparison methods, experimental results, and ablation study are described successively.

\begin{table*}[t]
\caption{\textbf{Details of three re-organized remote sensing semantic segmentation datasets in the FL setting.}}
\begin{tabular*}{\hsize}{@{}@{\extracolsep{\fill}}llll@{}}
\toprule 
                    & FedFBP             & FedCASID           & FedInria           \\  \midrule 
Source Dataset       & FBP~\cite{ISPRS_FBP}               & CASID~\cite{ISPRS_CASID}               & Inria~\cite{IGARSS_IAIL}                \\
Released Year       & 2023               & 2023               & 2017                \\
Task                & land cover mapping & land cover mapping & building extraction \\
Heterogeneity Scale of Data Islands & region-level   & climate zone-level          & city-level           \\
Number of Participants      & 6                  & 4                  & 5                   \\
Number of Annotated Classes   & 24                 & 4                  & 2                   \\
GSD (meter/pixel)       & 4                  & 1                  & 0.3                 \\
 Image Size (pixels) & 6800 $\times$ 7200          & 5000 $\times$ 5000          & 5000 $\times$ 5000           \\
Number of Original Images & 150          & 980          & 180           \\
Cropped Image Size  & 512 $\times$ 512            & 512 $\times$ 512            & 512 $\times$ 512        \\ 
Re-organized Data Size Ratio  & 13:37:10:36:40:14     & 209:202:284:284    & 36:36:36:36:36        \\ 
\bottomrule    
\end{tabular*}
\label{tab:datasets}
\end{table*}

\begin{figure*}[t]
    \centering
    \includegraphics[width=1\linewidth]{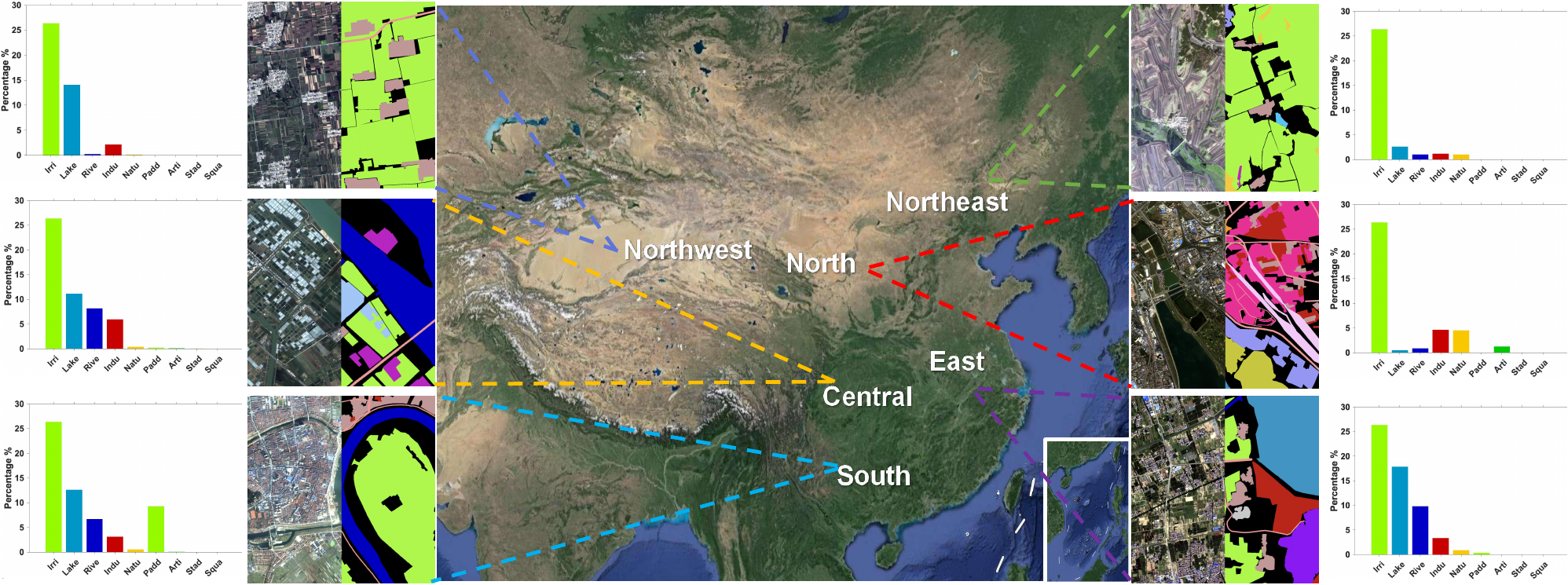}
    \caption{\textbf{Illustration of the geographic heterogeneity in the FedFBP dataset.} From the samples of satellite Images in various geographic regions, it is evident that geographic heterogeneity is common in remote sensing semantic segmentation tasks under distributed FL scenarios and is closely coupled. geographic heterogeneity manifests as both class-distribution heterogeneity and object-appearance heterogeneity.}
    \label{fig:FedFBP}
\end{figure*}

\subsection{Dataset Description and Evaluation Metrics}
To assess and measure the effectiveness of our GeoFed framework, We conduct experiments on three public datasets. They are FBP~\cite{ISPRS_FBP}, CASID~\cite{ISPRS_CASID} and Inria~\cite{IGARSS_IAIL}. We re-organize these datasets and construct three datasets (i.e., FedFBP, FedCASID, FedInria) in remote sensing scenarios under FL settings. The details of the three datasets are summarized in Table~\ref{tab:datasets} for clarity. We randomly separate each institution's data with a ratio of 6: 2: 2 into train, validation and test. Thus, each institution owns a local train, validation, and test dataset. Due to the complex category system of the FedFBP dataset, we aim to avoid the scenario where only a few classes dominate the test set in the distributed training setup. Therefore, we first cropped the data from each client into patches and then randomly split them into train, validation, and test sets. 

\textbf{(1) FedFBP Dataset.}
FedFBP is annotated according to a 24-category system and contains a total of 150 satellite images from China with a size of 6800 $\times$ 7200 pixels. The Ground Sampling Distance (GSD) is 4 meters. Based on the geographic divisions of China, we divide the dataset into six regions (i.e., Northeast, North, Northwest, Central, East, and South). In Fig~\ref{fig:FedFBP}, we illustrate the geographic heterogeneity reflected in the FedFBP dataset. It present the severe class-distribution heterogeneity. This fine-grained dataset exhibits a common and pronounced long-tail effect. Some institutions lack certain categories, resulting in inconsistencies in the classification system and exacerbating the issue of category heterogeneity. 

\begin{figure*}[t]
    \centering
    \includegraphics[width=1\linewidth]{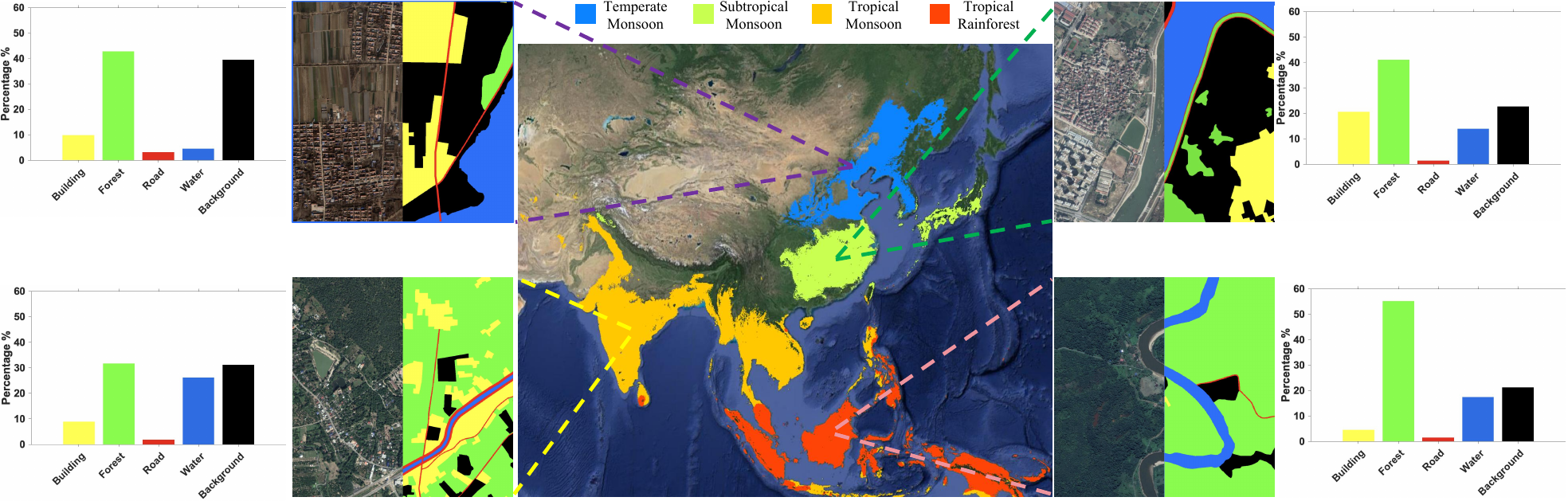}
    \caption{\textbf{Illustration of the geographic heterogeneity in the FedCASID dataset.} Samples above are collected from four institutions across climate zones, respectively.}
    \label{fig:FedCASID}
\end{figure*}

\textbf{(2) FedCASID Dataset.}
FedCASID contains a total of 980 images with a size of 5000 $\times$ 5000 pixels from four typical climatic zones (i.e., tropical rainforest, tropical monsoon, subtropical monsoon and temperate monsoon). Its GSD is 1 meter. Due to climate being an important factor in land cover features, there is a high geographic heterogeneity among the four institutions. This dataset is annotated with four categories: building, water, forest, and road. These categories exhibit noticeable differences due to variations in geographic regions. we illustrate the geographic heterogeneity reflected in the FedCASID dataset in Fig~\ref{fig:FedCASID}.

\begin{figure*}[t]
    \centering
    \includegraphics[width=1\linewidth]{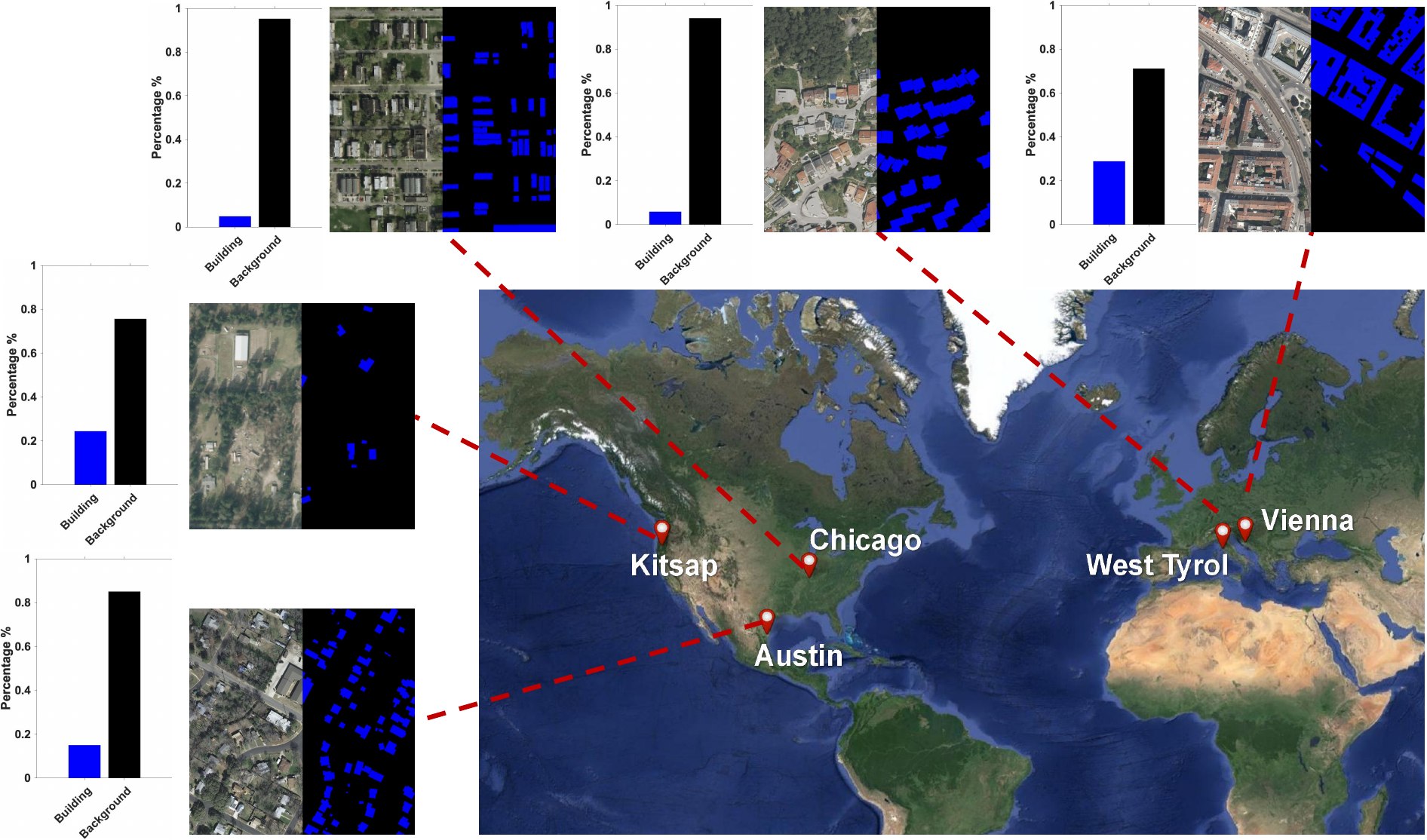}
    \caption{\textbf{Illustration of the geographic heterogeneity in the FedInria dataset.} The samples above are collected from five institutions across cities, respectively.}
    \label{fig:FedInria}
\end{figure*}

\textbf{(3) FedInria Dataset.}
FedInria is an aerial image dataset for building extraction with a spatial resolution of 0.3 m. It contains 180 images of size 5000 $\times$ 5000 pixels. The data is divided into five different cities (i.e., Austin, Chicago, Kitsap, West Tyrol, and Vienna), with significant heterogeneity in architectural styles between institutions. The heterogeneity in class distribution manifests in the building segmentation task as an imbalance between foreground and background, resulting from variations in the density of buildings. Because of the variation in economic development, climate and some other complex factors, the dataset was partitioned to reflect different regions, each with varying building densities and architectural styles. We illustrate the geographic heterogeneity reflected in the FedInria dataset in Fig~\ref{fig:FedInria}.

The mean intersection-over-union~(mIoU) metric is adopted to evaluate the segmentation performance on the FedFBP and FedCASID datasets. Additionally, the IoU metric is utilized to evaluate the performance on the FedInria dataset.

Local testing performance is assessed across different institutions to evaluate models' performance. This metric reflects the model's adaptability to local personalized data. By merging all test sets and conducting a global evaluation, the global test metric provides insights into the model's overall generalization capability. Specifically, for evaluation of local and global performance, we define as follows:

\begin{equation}
M^{local} = \dfrac{1}{n}\sum_{i=1}^{n} \left(\dfrac{1}{C}\sum \frac{TP_i}{TP_i + FP_i + FN_i}\right),
\label{eq:local_test}
\end{equation}

\begin{equation}
M^{global} =\dfrac{1}{C}\sum \frac{\sum_{i=1}^{n}TP_i}{\sum_{i=1}^{n}TP_i+\sum_{i=1}^{n}FP_i+\sum_{i=1}^{n}FN_i}.
\label{eq:global_test}
\end{equation}

Evaluation metrics are calculated by counting the pixels for false
negatives (FN), true positives (TP), and false positives (FP) in the i-th institution.

\subsection{Implementation Details}

The experiments simulate the FL process with PyTorch using a single NVIDIA RTX Titan GPU with 24-GB GPU memory in a serial manner. Due to memory limitations, all images are cropped into patches with a size of $512\times 512$ pixels. The background pixels of the FedFBP and FedCASID datasets are ignored during training. Unless otherwise specified, all experiments use DeepLabv3+~\cite{ECCV_DeepLabV3p} as the segmentation model. The ResNet-50~\cite{CVPR_ResNet} pretrained on ImageNet~\cite{CVPR_ImageNet} is used as the backbone. The optimizer is SGD with an initial learning rate of $\eta_l=0.001$, and the batch size is set to 4. We employ a poly learning rate scheduler. The server performs $T=100$ aggregation operations to ensure convergence, and institutions update themselves with local training epoch $E=1$. The balancing network adopts the ResNet-18. In the experiments, we set $\lambda_{1} = 0.4$, $\lambda_{2} = 0.6$, the temperature $\tau = 0.05$, and the update ratio of essential features $\gamma = 0.8$.

\subsection{Comparison Methods}
Based on three paradigms of remote sensing semantic segmentation (i.e., Local learning, Centralized learning and our novel Federated learning), we compare and validate our GeoFed framework for bridging remote sensing data islands. We also assess the effectiveness of our geographic heterogeneity-aware design tailored for remote sensing data, which is superior to the existing FL methods. A brief summary of these three paradigms is provided sequentially below.
\subsubsection{Local Learning (LL)}
In the dilemma of remote sensing data islands, institutions focus more on their own data. Training samples are relatively scarce and homogeneous. LL refers to each institution solely using its private data for traditional deep learning training. In this way, each institution can obtain a model exclusive to its own region. However, due to the inability to access data from other institutions, such models lack collaborative learning with external regional knowledge, making them prone to overfitting and resulting in relatively poor global generalization. This approach represents a real scenario where each institution independently trains its model and produces local LULC mapping products, eventually merging all results to obtain a large-scale remote sensing semantic segmentation map.

\subsubsection{Centralized Learning (CL)}
In the traditional deep learning paradigm, to train a well-generalized model, various remote sensing data are typically aggregated onto the same device for CL. This approach benefits from the large volume of data, leading to improved generalization. Meanwhile, it overlooks the advantages of zoning mapping in remote sensing interpretation. Within each geographic zone, the characteristics are relatively homogeneous, while there is heterogeneity between geographic regions. This implies that the same region may follow relatively consistent geographic patterns. The optimization goal in centralized training is a global loss, lacking the exploration of knowledge about the geographic characteristics within each geographic region. It should be noted that this approach violates data privacy assumptions. However, it can be anticipated that this provides a potential performance upper limit reference for FL. Our goal is to approach or even surpass this reference as closely as possible.

\subsubsection{Federated Learning (FL) frameworks}
For comparison with existing state-of-the-art methods, We choose FedAvg, FedProx, Moon, FedDisco and Elastic. A brief summary of these FL methods is provided below. Here, FedAvg with a naive aggregation serves as the baseline method.
\begin{itemize}
    \item \textbf{FedAvg}~\cite{AISTATS_FedAvg}: It is a practical method for the federated learning of deep networks based on iterative model averaging.
    \item \textbf{FedProx}~\cite{MLSys_FedProx}: It introduces a proximal term loss to constrain the deviation in local updates. However, this model parameter-wise constraint does not consider the heterogeneity characteristics of the data, which may result in slow convergence speed.
    \item \textbf{Moon}~\cite{CVPR_Moon}: Its key idea is to utilize the similarity between model representations to correct the local training of individual parties, i.e., conducting contrastive learning at the model level.
    \item \textbf{FedDisco}~\cite{ICML_FedDisco}: Its aggregation weights not only involve both the dataset size and the discrepancy value but also contribute to a tighter theoretical upper bound of the optimization error.
    \item \textbf{Elastic}~\cite{CVPR_Elastic}: Elastic reduces the magnitudes of updates to the more sensitive parameters to prevent the server model from drifting to local distribution and, conversely, boosts updates to the less sensitive parameters to explore different local distributions better. 
\end{itemize}

\subsection{Comparison with State-of-the-Arts}
In this section, we sequentially compare our GeoFed framework with previous state-of-the-art methods on three datasets. We report the region-wise performance of our method and other methods in Table~\ref{tab: comparison_with_sota_FedFBP}, \ref{tab: comparison_with_sota_FedCASID} and \ref{tab: comparison_with_sota_FedInria}. To provide a clearer comparison of the experiments, we present the mIoU results of all comparison methods on the local test sets of each institution. The "Average" is calculated by~\eqref{eq:local_test}, which indicates the mean result of each method across all local test sets, while "Global" is calculated by~\eqref{eq:global_test}, which represents the global mIoU result after merging the test sets. 
\subsubsection{Experiments on the FedFBP Dataset}
\begin{table*}[htbp]
\centering
\caption{\textbf{Comparisons with previous state-of-the-art methods under the mIoU (\%) metric on FedFBP datasets.} Each row represents the performance of various methods in the corresponding institution test set of that row. The best results are marked in \textbf{bold}. $^\dagger$ denotes centralized learning, which serves as an upper bound but can not be realized because of data islands actually. The same applies to the subsequent tables.}
\begin{tabular*}{\hsize}{@{}@{\extracolsep{\fill}}cc|c|ccccccc@{}}
\bottomrule	
\multicolumn{2}{c|}{}          & CL$^\dagger$   & LL  & FedAvg~\cite{AISTATS_FedAvg}      & FedProx~\cite{MLSys_FedProx}   & Moon~\cite{CVPR_Moon}   & FedDisco~\cite{ICML_FedDisco}   & Elastic~\cite{CVPR_Elastic}  & GeoFed (Ours) \\
\hline
\multirow{7}{*}{FedFBP}   & Northeast  & 55.34   &   44.39     & 49.89       & 47.82     & 49.71    & 50.17     & 50.08    & \textbf{52.41}      \\
                       & North      &  64.93 &    57.59     & 61.32       & 60.42     & 61.47    & 61.39     & 61.65    & \textbf{63.52}      \\
                       & Northwest  &  49.13 &   40.68      & 46.28       & 45.70     & 45.35    & 46.80     & 46.80    & \textbf{48.54}      \\
                       & East       & 64.13  &    56.38     & 60.50       & 60.29     & 60.22    & 59.66     & 60.74    & \textbf{61.62}      \\
                       & Central    &  65.34 &   56.88      & 60.38       & 59.82     & 60.70    & 59.65     & 61.43    & \textbf{62.41}      \\
                       & South      &  63.73 &   55.38      & 60.40       & 59.28     & 60.72    & 59.67     & 60.45    & \textbf{62.37}      \\
                       \cline{2-10}
                       & Average    & 60.43  &  51.88       & 56.46       & 55.55     & 56.36    & 56.22     & 56.86    & \textbf{58.48}      \\
                        
                       & Global    & 68.76  &  59.83       & 63.13       & 62.92     & 63.23    & 63.31     & 64.20    & \textbf{66.70}      \\

                      \toprule
\end{tabular*}
  \label{tab: comparison_with_sota_FedFBP}
\end{table*}

\begin{table*}[htbp]
\caption{\textbf{Comparisons for each category's IoU (\%) scores on the FedFBP dataset.} Notes: Indu (industrial area), Urba (urban residential), Rura (rural residential), Stad (stadium), Squa (square), Over (overpass), Rail (railway station), Airp (airport), Padd (paddy field), Irri (irrigated field), Dryc (dry cropland), Gard (garden land), Arbo (arbor forest), Shru (shrub forest), Natu (natural meadow), Arti (artificial meadow), Rive (river), Fish (fish pond), Bare (bare land).}
\label{tab:class_FBP}
\begin{tabular*}{\hsize}{@{}@{\extracolsep{\fill}}llllllllllllll@{}}
\toprule
       & \textbf{mIoU} & \textbf{IoU:} & Indu & Padd & Irri & Dryc & Grad & Arbo & Shru & Park & Natu & Arti & Rive \\  \cline{4-14}
CL$^\dagger$           & 68.76 &       & 74.46 & 67.28 & 94.45 & 82.48 & 50.96 & 95.26 & 89.66 & 47.25 & 77.94 & 56.68 & 83.42 \\ \midrule
LL           & 59.83 &       & 70.47 & 46.71 & 89.85 & 75.70 & 48.10 & 91.55 & 86.94 & 34.48 & 68.99 & 53.37 & 73.73 \\
FedAvg~\cite{AISTATS_FedAvg}       & 63.13 &       & 71.01 & 53.24 & 90.87 & 79.28 & 47.63 & 91.96 & 87.35 & 34.89 & 74.12 & 52.05 & 78.12 \\
FedProx~\cite{MLSys_FedProx}      & 62.92 &       & 71.18 & 56.31 & 90.39 & 78.33 & 47.49 & 91.48 & 86.98 & 38.27 & 73.18 & 52.22 & 78.68 \\
Moon~\cite{CVPR_Moon}         & 63.23 &       & 70.92 & 54.34 & 90.64 & 78.61 & 47.88 & 91.95 & 87.03 & 35.61 & 72.93 & 51.55 & 78.70 \\
FedDisco~\cite{ICML_FedDisco}     & 63.31 &       & 71.08 & 55.62 & 90.43 & 78.66 & 45.91 & 91.56 & 87.02 & 37.99 & 73.06 & 50.81 & 76.47 \\
Elastic~\cite{CVPR_Elastic}      & 64.20 &       & 71.27 & 61.80 & 90.84 & 79.64 & 47.01 & 92.05 & 87.35 & 37.92 & 73.60 & 52.35 & 79.24 \\
GeoFed~(Ours) & \textbf{66.70} &       & \textbf{73.12} & \textbf{65.09} & \textbf{93.15} & \textbf{82.45} & \textbf{49.08} & \textbf{94.05} & \textbf{89.39} & \textbf{40.34} & \textbf{76.86} & \textbf{54.72} & \textbf{81.69} \\ \midrule
             & Urba & Lake & Pond & Fish & Snow & Bare & Rura & Stad & Squa & Road & Over & Rail & Airp \\   \cline{2-14}
CL$^\dagger$ & 79.90 & 91.69 & 43.44 & 55.98 & 57.09 & 83.71 & 81.24 & 65.08 & 26.26 & 69.56 & 68.36 & 43.28 & 64.91 \\ \midrule
LL           & 76.15 & 86.39 & 36.06 & 36.53 & 42.89 & 80.22 & 77.35 & 47.63 & 5.95  & 63.28 & 58.69 & 32.84 & 51.96 \\
FedAvg~\cite{AISTATS_FedAvg}       & 76.66 & 87.70 & 36.09 & 44.65 & 45.31 & 84.16 & 77.73 & 57.02 & 16.55 & 65.31 & 62.92 & 38.56 & 61.94 \\
FedProx~\cite{MLSys_FedProx}      & 76.61 & 87.61 & 35.50 & 45.24 & 47.74 & 81.86 & 77.48 & 55.79 & 12.99 & 64.79 & 61.85 & 37.02 & 60.96 \\
Moon~\cite{CVPR_Moon}         & 76.60 & 87.99 & 36.07 & 45.03 & 47.04 & 83.49 & 77.70 & 58.48 & 17.77 & 65.24 & 62.15 & 36.93 & 62.92 \\
FedDisco~\cite{ICML_FedDisco}     & 76.52 & 86.90 & 34.56 & 47.51 & 47.82 & 82.95 & 77.68 & 57.32 & 19.18 & 65.22 & 62.28 & 39.17 & 60.70 \\
Elastic~\cite{CVPR_Elastic}      & 76.83 & 88.10 & 36.42 & 51.26 & 47.80 & 83.33 & 77.79 & 58.55 & 18.16 & 65.51 & 62.86 & 38.19 & 62.00 \\
GeoFed~(Ours) & \textbf{78.82} & \textbf{90.47} & \textbf{38.71} & \textbf{54.79} & \textbf{48.81} & \textbf{86.13} & \textbf{79.94} & \textbf{61.29} & \textbf{23.14} & \textbf{67.93} & \textbf{65.92} & \textbf{41.55} & \textbf{63.27} \\
\bottomrule
\end{tabular*}
\end{table*}
\begin{table*}[htbp]
\caption{\textbf{Cross-institution test results for the performance of the six regions in the FedFBP dataset.} Note that the underlined results correspond to the LL method in~Tab.\ref{tab: comparison_with_sota_FedFBP}}.
\begin{tabular*}{\hsize}{@{}@{\extracolsep{\fill}}l|llllll|l@{}}
\bottomrule
\diagbox[width=8em]{Train}{Test}       & Northeast      & North          & Northwest      & East           & Central        & South          & Average        \\
\hline
Northeast & \underline{44.39}          & 33.45          & 42.84          & 46.56          & 43.50          & 34.44          & 41.26          \\
North    & 39.54          & \underline{57.59}          & 32.01          & 35.89          & 40.77          & 35.82          & 40.21          \\
Northwest & 40.18          & 38.77          & \underline{40.68}          & 41.36          & 40.56          & 40.25          & 40.78          \\
East      & 46.32          & 35.62          & 43.79          & \underline{56.38}          & 42.18          & 37.46          & 43.74          \\
Central   & 43.76          & 39.98          & 36.08          & 40.40          & \underline{56.88}          & 32.57          & 41.72          \\
South    & 37.51          & 37.54          & 41.22          & 42.80          & 37.84          & \underline{55.38}          & 42.22          \\
\hline
GeoFed (Ours)         & \textbf{52.41} & \textbf{63.52} & \textbf{48.54} & \textbf{61.62} & \textbf{62.41} & \textbf{62.37} & \textbf{58.48}\\
\toprule
\end{tabular*}
\label{tab:domain_gap}
\end{table*}

Six institutions from different regions of China participate in the FedFBP dataset setting. From the mIoU comparison shown in Table~\ref{tab: comparison_with_sota_FedFBP}, We observe that our proposed GeoFed framework achieves state-of-the-art performance on three datasets and is very close to the results of centralized training. Our method surpasses the LL method by 6.87\%, demonstrating the advantages of FL and the effective aggregation of knowledge from different regions. Additionally, the gap between our method and the CL method is 2.06\% from the global test. In addition, our GeoFed attains an improvement of 1.62\% and 2.5\% on the local test and global test, respectively, compared to state-of-the-art methods that are not designed for geographic heterogeneity. 
 
 Subsequently, we illustrate the IoU of all methods to compare the class-wise performance in Table~\ref{tab:class_FBP}. In comparison, we illustrate the IoU performance of classes for several methods in Table~\ref{tab:class_FBP}. Our approach achieves performance improvements in almost every class, especially some tail classes (e.g., stadium, square and natural meadow).
 
\begin{figure*}[htbp]
    \centering
    \includegraphics[width=0.9\linewidth]{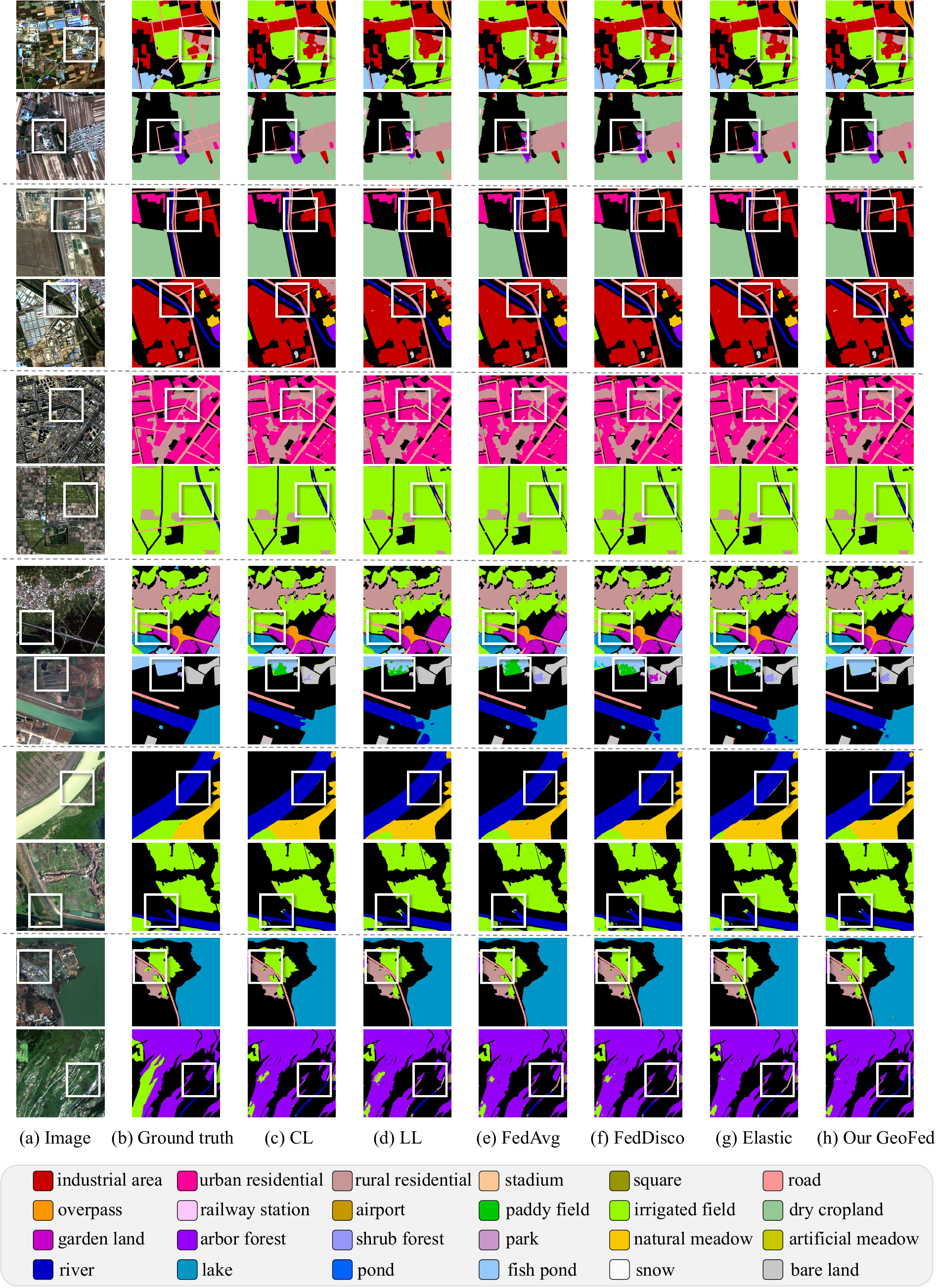}
    \caption{\textbf{Visualization of the comparison among state-of-the-art methods on FedFBP.} The samples from the first to the last rows are collected from the Northeast, North, Northwest, Central, East, and South regions, respectively. It is best viewed in colors.}
    \label{fig:FedFBP_vis}
\end{figure*}

In Fig~\ref{fig:FedFBP_vis}, we present two patch results for the test set of each regional institution. Large-size image results are not shown since we adopted the strategy of patching the data before splitting it into datasets, as mentioned before. From the visualization results, it is clear that our GeoFed is more capable of interpreting the land cover categories. Competitor methods are affected more by geographic heterogeneity, resulting in poor performance relatively.

We further implement cross-institution tests on the FedFBP dataset to verify the superiority of our GeoFed in bridging data islands. Table~\ref{tab:domain_gap} presents the results of different institutions using private datasets for local model training and testing on other test sets. It can be seen that the geographic heterogeneity of different institutions leads to a severe institution gap, resulting in poor generalization performance of the models. However, our GeoFed framework effectively enables communication among individual models and balances erudite and expertise well. Compared to each locally trained model, which performs well in its own region, overall performance is improved by an average of about 15\%. Our GeoFed successfully alleviates overfitting in each local model.

\subsubsection{Experiments on the FedCASID Dataset}
\begin{table*}[t]
\centering
\caption{\textbf{Comparisons with previous state-of-the-art methods under the mIoU (\%) metric on the FedCASID datasets.} Each row represents the test results of various methods in the corresponding institution test dataset of that row. The abbreviations for regions are defined as: (TroRf: tropical rainforest, TroMs: tropical monsoon, SubMs: subtropical monsoon, TemMs: temperate monsoon).}
\begin{tabular*}{\hsize}{@{}@{\extracolsep{\fill}}cc|c|ccccccc@{}}
\bottomrule	
\multicolumn{2}{c|}{}          & CL$^\dagger$   & LL  & FedAvg~\cite{AISTATS_FedAvg}      & FedProx~\cite{MLSys_FedProx}   & Moon~\cite{CVPR_Moon}   & FedDisco~\cite{ICML_FedDisco}   & Elastic~\cite{CVPR_Elastic}  & GeoFed (Ours) \\
\hline
\multirow{5}{*}{FedCASID} &TroRf    & 70.78  &   60.88      & 63.42& 63.37& 58.96& 61.30& 64.93& \textbf{66.21}\\
                          &TroMs    & 78.01  &   65.52      & 74.96& 74.17& 72.16& 75.41& 75.15& \textbf{78.90}\\
                          &SubMs    & 79.16  &   76.35      & 76.06& 77.08& 74.62& 76.05& 77.06& \textbf{79.39}\\
                          &TemMs    & 69.15  &   58.01      & 65.03& 65.63& 63.25& 66.91& 66.56& \textbf{68.82}\\
                        \cline{2-10}
                          &Average  & 74.28  &   64.19      & 69.87&70.06& 67.25& 69.92& 70.93& \textbf{73.33}\\ 
                          & Global  & 75.40  &   61.31      & 71.18& 71.30& 68.80& 70.95& 72.17& \textbf{74.34}\\ \toprule
\end{tabular*}
  \label{tab: comparison_with_sota_FedCASID}
\end{table*}

\begin{table*}[t]
\caption{\textbf{Comparison of the IoU (\%) performance of classes on the FedCASID dataset.}}
\begin{tabular*}{\hsize}{@{}@{\extracolsep{\fill}}lllllllllll@{}}
\bottomrule
\multirow{2}{*}{Method} & \multicolumn{4}{c}{Local}        &      & \multicolumn{4}{c}{Global}       &      \\  \cline{2-5}  \cline{7-10}
                        & building & Road & Forest & Water & mIoU & Building & Road & Forest & Water & mIoU \\
                        \cline{1-5}  \cline{7-10}
CL$^\dagger$                      &83.93          &46.18      &95.95        &71.05       &74.28      &85.61          &47.69      &95.96        &72.35       &75.40      \\ \midrule   
LL                      &76.60          &45.65      &91.50        &43.01       &64.19      &76.24          &37.89      &91.34        &39.78       &61.31      \\
FedAvg~\cite{AISTATS_FedAvg}                  &80.61          &42.57      &94.22        &62.08       &69.87      &82.63          &43.47      &94.17        &64.44       &71.18      \\
FedProx~\cite{MLSys_FedProx}                 &79.26          &43.50      &94.06        &63.42       &70.06      &78.44          &36.09      &93.05        &45.60       &63.30      \\
Moon~\cite{CVPR_Moon}                    &79.56          &40.36      &94.49        &54.59       &67.25      &81.77          &41.56      &94.41        &57.44       &68.80      \\
FedDisco~\cite{ICML_FedDisco}                &81.99          &42.35      &94.92        &60.42       &69.92      &83.90          &43.59      &94.83        &61.50       &70.95      \\
Elastic~\cite{CVPR_Elastic}                 &82.01          &43.49      &95.33        &66.87       &70.93      &83.86          &45.01      &95.32        &68.51       &72.17      \\
GeoFed (Ours)           &\textbf{84.70}          &\textbf{45.87}      &\textbf{95.34}        &\textbf{67.41}       &\textbf{73.33}      &\textbf{84.68}          &\textbf{47.42}      &\textbf{95.44}        &\textbf{69.80}       &\textbf{74.34}  \\ \toprule \end{tabular*}
  \label{tab: class_IoU_FedCASID}
\end{table*}

\begin{figure*}[htbp]
    \centering
    \includegraphics[width=0.9\linewidth]{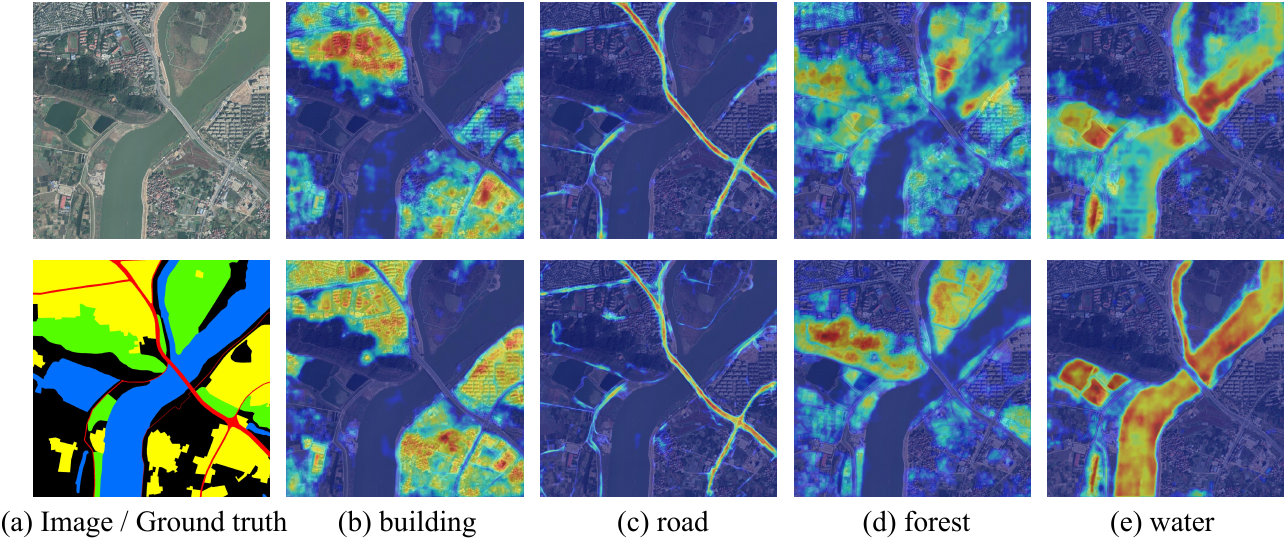}
    \caption{\textbf{The Grad-CAM visualization.} The first row is calculated with Elastic and the second row is calculated with our GeoFed framework.}
    \label{fig:cam}
\end{figure*}

Table~\ref{tab: comparison_with_sota_FedCASID} illustrates the comparisons with previous state-of-the-art methods on the FedCASID datasets. The models trained using CL, LL, and FL methods were locally tested on the test sets from tropical rainforest, tropical monsoon, subtropical monsoon, and temperate monsoon regions, and globally tested on the merged dataset. From the local test results across different regions, it is evident that the tropical monsoon and subtropical monsoon regions achieved relatively higher mIoU results, approximately 10\% higher on average compared to the other two regions. The superior local test mIoU performance in these two regions is primarily attributed to the higher IoU for buildings, roads, and water bodies. We consider that in the tropical rainforest region, the dense forest coverage causes buildings, roads, and water bodies to be frequently obscured by the forest, leading to challenges in accurately identifying their boundaries. In the temperate monsoon region, water bodies are fewer and more fragmented. These potential factors contribute to the increased difficulty of local recognition in these areas, resulting in overall lower mIoU performance.

\begin{figure*}[htbp]
    \centering
    \includegraphics[width=0.9\linewidth]{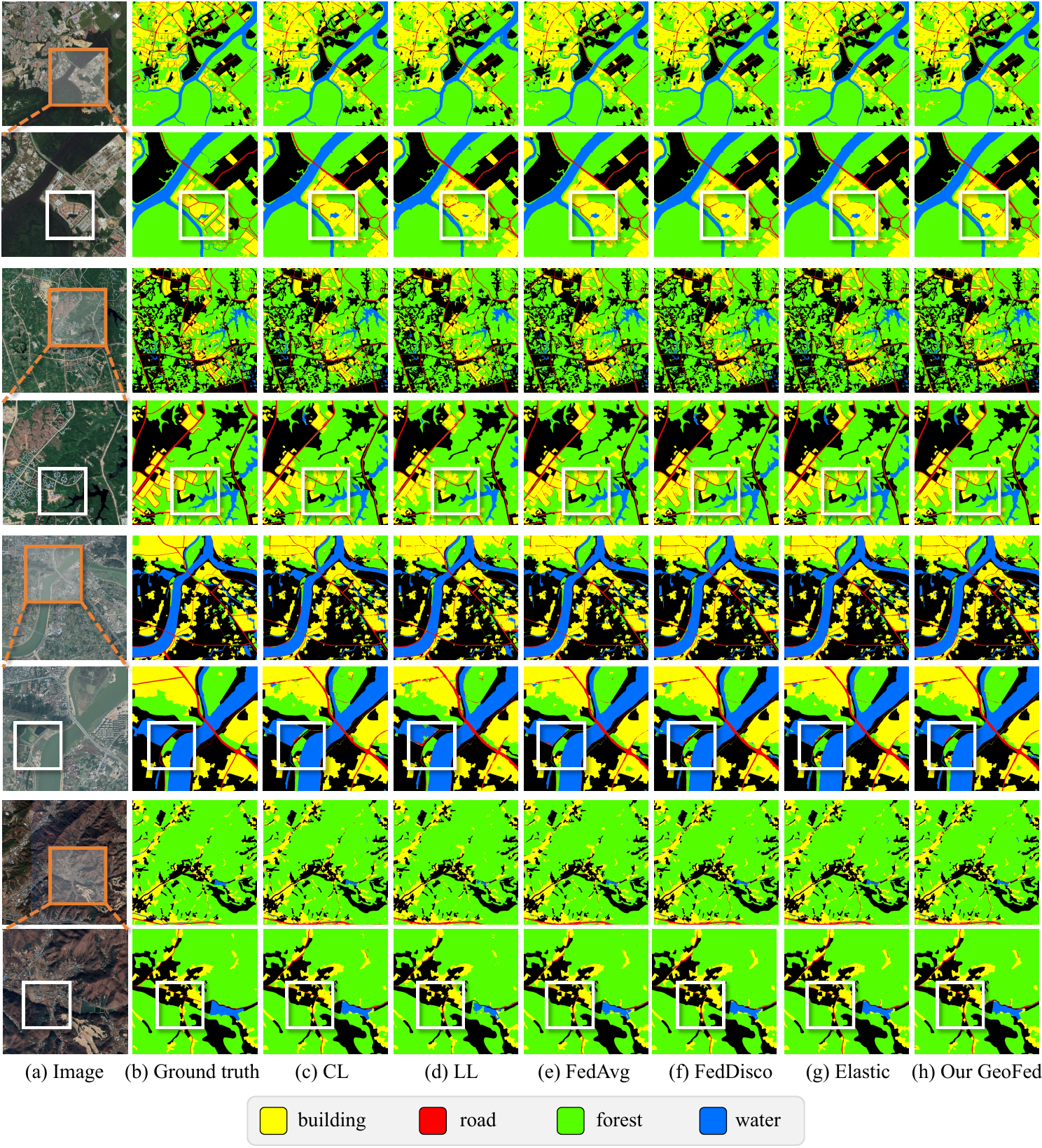}
    \caption{\textbf{Visualization of the comparison among state-of-the-art methods on FedCASID.} The samples from the first to the last rows are collected from the tropical rainforest, tropical monsoon, subtropical monsoon and temperate monsoon regions, respectively.}
    \label{fig:FedCASID_vis}
\end{figure*}

\begin{table*}[htbp]
\centering
\caption{\textbf{Comparisons with previous state-of-the-art methods under the mIoU (\%) metric on FedInria datasets.} Each row represents the performance of various methods in the corresponding institution test set of that row.}
\begin{tabular*}{\hsize}{@{}@{\extracolsep{\fill}}cc|c|ccccccc@{}}
\bottomrule	
\multicolumn{2}{c|}{}          & CL$^\dagger$   & LL  & FedAvg~\cite{AISTATS_FedAvg}      & FedProx~\cite{MLSys_FedProx}   & Moon~\cite{CVPR_Moon}   & FedDisco~\cite{ICML_FedDisco}   & Elastic~\cite{CVPR_Elastic}  & GeoFed (Ours) \\
\hline
\multirow{6}{*}{FedInria}  & Austin     &  68.32  &   63.12 & 65.27       & 64.35     & 66.71    & 66.41     & 65.21    & \textbf{67.16}      \\
                       & Chicago    &65.87   &  62.17       & 63.56       & 63.41     & 64.02    & \textbf{64.93}     & 63.37    & 64.84      \\
                       & Kitsap     & 57.24  &   55.42      & 54.42       & 54.02     & 55.31    & 56.58     & 55.51    & \textbf{56.31}      \\
                       & West Tyrol &68.56   &  65.34       & 64.76       & 64.92     & 65.23    & 65.67     & 65.34    & \textbf{67.97}      \\
                       & Vienna     &74.33   &  71.27       & 72.88       & 73.47     & 73.22    & 73.91     & 73.70    & \textbf{73.99}      \\
                        \cline{2-10}
                       & Average    &66.86   &   63.46      & 64.18       & 64.03     & 64.90    & 65.50     & 64.63    & \textbf{66.05}  \\ 
                        & Global    &69.07   &   43.10      & 65.27       & 65.11     & 64.53    & 65.76     & 64.89    & \textbf{65.92}   
                      \\ \toprule
\end{tabular*}
  \label{tab: comparison_with_sota_FedInria}
\end{table*}

\begin{figure*}[htbp]
    \centering
    \includegraphics[width=0.9\linewidth]{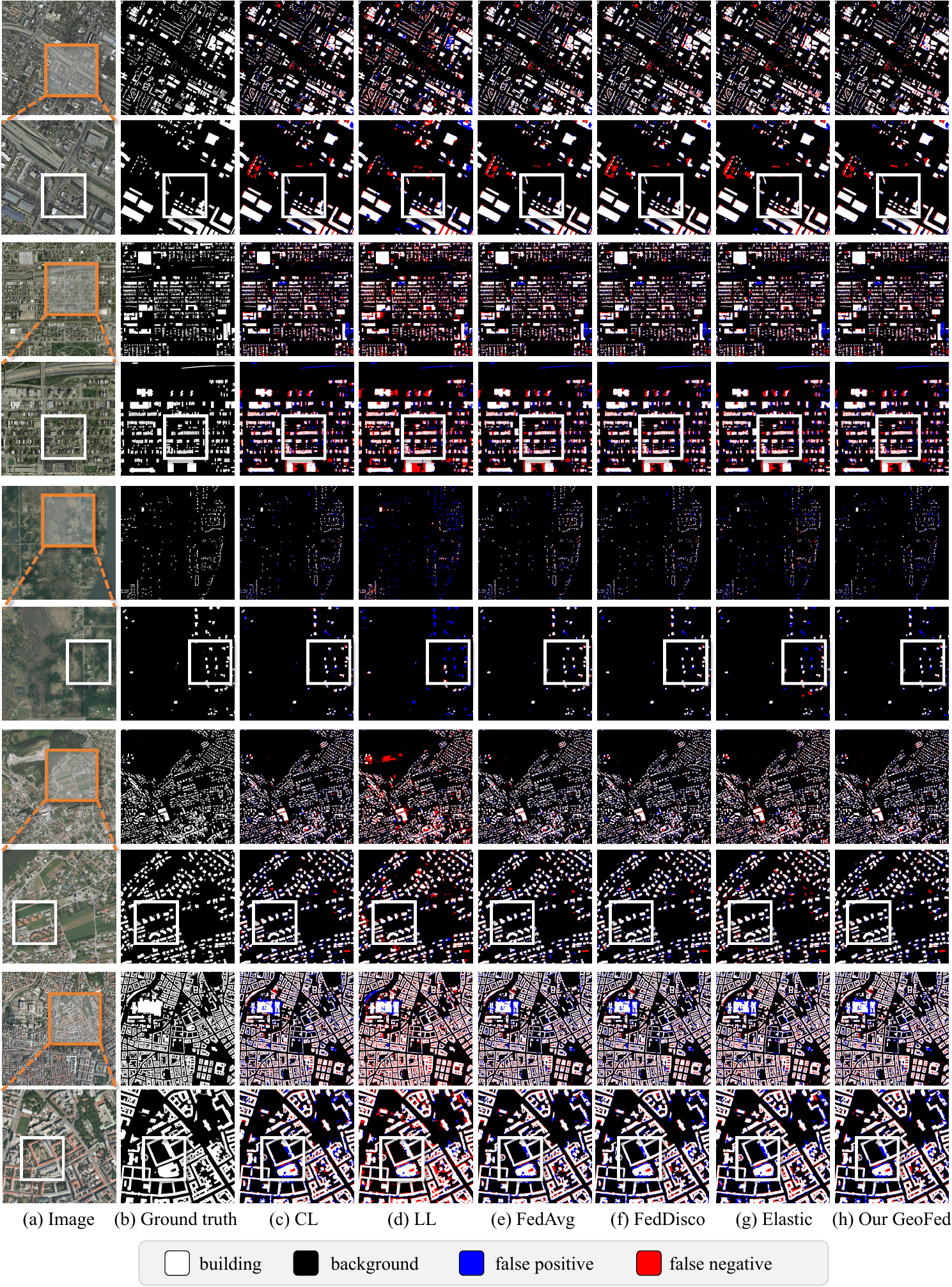}
    \caption{\textbf{Visualization of the comparison among state-of-the-art methods on FedInria.} The samples from the first to the last rows are collected from the Austin, Chicago, Kitsap, West Tyrol, and Vienna regions, respectively.}
    \label{fig:FedInria_vis}
\end{figure*}

It is clearly observed that the CL method, due to its access to all institutions' data (though impractical in real-world scenarios), achieves relatively higher performance across all test sets. In the FedCASID dataset, the local testing performance of the LL method shows relatively smaller gaps compared to other methods. This is primarily because the FedCASID dataset is large, and each institution possesses a relatively substantial amount of data. Therefore, each institution can achieve relatively high performance by overfitting to their local private data. This overfitting leads to the similar low generalization performance presented in Table~\ref{tab:domain_gap}.

The geographic heterogeneity brought about by different climates, such as the completely different types of forests in tropical rainforest and temperate monsoon regions, results in biases during the local update process of models at each institution. The dual impact of heterogeneous class distribution and target appearance leads to performance degradation of FedAvg, FedProx, Moon, FedDisco, and Elastic to varying degrees. FedProx, in particular, suffers a performance decline because it only regularizes parameters at the model parameter level without feature-level constraints. Moon improves model performance relative to the previous round of updates, but the local performance of some clients remains insufficient due to the lack of methods balancing global and local updates.

Our GeoFed method consistently exhibits optimal performance in both local and global tests. By extending global information, it prevents each institution from overly focusing on local information during training, introducing global class information in the process. Simultaneously, through intrinsic feature mining, different target appearances, such as forests, water bodies, buildings and roads, are subject to consistency constraints during local training, enabling the model to learn intrinsic rather than biased local features. Finally, the global-local balance module effectively combines global and local information during the training process.

Fig~\ref{fig:cam} illustrates the Gradient-weighted Class Activation Mapping (Grad CAM)~\cite{ICCV_GradCAM}. Our GeoFed focuses more accurately on the corresponding land cover category, which shows superior feature learning capabilities.

In Table~\ref{tab: class_IoU_FedCASID}, we report the class-wise metric performance of each method in both global and local tests. It is evident that roads, which are relatively sparse and typically exhibit a narrow, elongated structure, are more challenging to segment, resulting in lower IoU scores. In both local and global tests, our GeoFed framework achieves higher IoU scores across all four categories. This indicates that GeoFed can more effectively aggregate knowledge from various institutions, mitigating the impact of geographic heterogeneity. Notably, it achieved a 2.38\% and 2.41\% improvement compared to the second-best Elastic method in IoU for the road category in local and global tests, respectively.

\begin{table}[t]
  \centering
  \setlength{\tabcolsep}{2mm}
    \caption{\textbf{Ablation study results of the proposed components.}}
  \begin{tabular}{cccccc}
    \toprule
    Method & GIE&EFM&LoGo &mIoU(\%)&$\Delta$(\%) \\
    \midrule
    Baseline &  & & &63.13& \\
    I & \checkmark & &&65.10&+1.97 \\
    II & &\checkmark & &64.57& +1.44\\
    III & & &\checkmark &64.87&+1.74 \\
      IV & \checkmark  &\checkmark &&65.15&+2.02 \\
     V & \checkmark  & &\checkmark&65.67&+2.54 \\
     VI &  &\checkmark  &\checkmark&66.14&+3.01 \\
     VII & \checkmark &\checkmark &\checkmark &\textbf{66.70}&+3.57 \\
    \bottomrule
  \end{tabular}
  \label{tab:ablation studies}
\end{table}

\subsubsection{Experiments on the FedInria Dataset}

In the experiment with the FedInria dataset, our GeoFed attains an improvement of 1.03\% under the IoU metric compared with the second-best model, Elastic. The comparison with the state-of-the-art methods is shown in Table~\ref{tab: comparison_with_sota_FedInria}.
Note that the FedIniria dataset is a building extraction task that only involves the building and background classes. Due to the geographic heterogeneity issues in the FedFBP datasets, which involve multi-class semantic segmentation, the challenge posed is more severe than FedInria, resulting in larger gaps. As a dataset with a single target category, the impact of geographic heterogeneity on FL is relatively minor in this context. Nevertheless, our method still achieved a slight performance improvement.
We observe that our method achieves stable performance improvements in almost every city. Some cities (e.g., Austin) with poor performance in LL achieve a significant performance improvement of 4.04\% through our method.

The semantic segmentation results are shown in Fig~\ref{fig:FedInria_vis}. Overall, our proposed GeoFed can extract building objects with much fewer commission errors (false positive) and omission errors (false negative). From the zoomed-in regions that are presented in the white boxes, it is apparent that our GeoFed preserves the edge more and can extract more accurate buildings.

\subsection{Ablation Study}

\subsubsection{Effectiveness of Our Proposed Components}

In Table~\ref{tab:ablation studies}, we conduct ablation experiments on the FedFBP dataset. We use FedAvg as the baseline method, where the GIE, EFM, and LoGo modules are used individually, resulting in mIoU improvements of 1.97\%, 1.44\%, and 1.74\%, respectively. This demonstrates the effectiveness of our proposed method in addressing heterogeneity issues. To alleviate class-distribution heterogeneity, the GIE module, and to mitigate object-appearance heterogeneity, the EFM module can achieve a performance improvement of 2.02\% when incorporated together. This demonstrates that these two issues need to be addressed collaboratively in the context of mixed geographic heterogeneity. The introduction of the LoGo module further enhances performance by balancing global and local knowledge, thereby complementing the other two modules. When all the modules we proposed are added to the baseline, our GeoFed achieves a performance improvement of 3.57\%. 

\subsubsection{Comparison of Semantic Segmentation Models}
Our proposed method is applicable to mainstream semantic segmentation models. To further validate the robustness of our proposed GeoFed, we conducted experiments using different models (i.e., OCRNet with a HRNet-W48 backbone and SegFormer with a MiT-B2 backbone) in the context of FL in Table~\ref{tab:models}. Between the HRNet-W48 and SegFormer-B2 models, our method benefits from stronger feature extraction backbones, resulting in performance improvements of 1.87\% and 2.94\%, respectively.

\begin{table*}[htbp]
\caption{\textbf{Scaling-up capability with the increasing of participanting institutions.}}
\centering
\label{tab:incres}
\begin{tabular}{cccccccccc}
\toprule
{Northeast} &
  {North} &
  {Northwest} &
  {East} &
  {South} &
  {Central} &
  {Local} &
  {$\Delta_L$} &
  {Global} &
  {$\Delta_G$} \\
  \midrule
\checkmark & & & & & &                                                         43.64 & &46.74 & \\
\checkmark & \checkmark & & & & &                                              47.53 &+3.89 &57.33 &+10.59 \\
\checkmark & \checkmark & \checkmark & & & &                                   49.25 &+5.61 &58.14 &+11.40\\
\checkmark & \checkmark & \checkmark & \checkmark & & &                        52.21 &+8.57 &61.41 &+14.67 \\
\checkmark & \checkmark & \checkmark & \checkmark & \checkmark & &             56.32 &+12.68 &65.34 &+18.60 \\
\checkmark & \checkmark & \checkmark & \checkmark & \checkmark & \checkmark &  58.48 &+14.84 &66.70 &+19.96
  \\ \bottomrule
\end{tabular}
\end{table*}

\begin{table}[htbp]
\centering
  \caption{\textbf{Results of implementing other semantic segmentation models.}}
 \begin{tabular}{ccc}
\toprule
Method  & HRNet-W48 + OCRNet     & MiT-B2 + SegFormer   \\
\midrule
FedAvg  & 67.01          & 67.24          \\
FedProx  & 67.12          & 67.90          \\
Moon   & 67.77          & 68.55          \\
FedDisco & 67.94          & 67.01          \\
Elastic& 67.01          & 68.11          \\
GeoFed (Ours)    & \textbf{68.57} & \textbf{69.64}\\
\bottomrule
\end{tabular}
  \label{tab:models}
\end{table}

\subsubsection{Scaling-up Capability with More Institutions}
To verify the effectiveness of the proposed GeoFed framework in facilitating collaboration among institutions and to evaluate the performance variation of the collaborative training model as more institutions join, we gradually introduce additional collaborating institutions into the GeoFed training process. The experimental results are shown in Table~\ref{tab:incres}. This experiment hints at the unseen institution's generalization ability of our GeoFed framework. The experimental results show that as more institutions, especially those with large data volumes, join the GeoFed distributed framework, significant improvements are achieved in both global and local performance. This indicates that our framework effectively helps bridge remote sensing data islands and fosters collaborative, mutually beneficial relationships among institutions.

\newpage
\section{Conclusion and future work}\label{conclusion}

This paper proposes a novel remote sensing semantic segmentation-oriented GeoFed framework to bridge these data islands and enable collaborative training of isolated data, considering the unique characteristics of remote sensing data. It includes multiple meticulously designed modules to address geographic heterogeneity. These modules are interconnected and optimized together under an overall loss function. More specifically, GIE aligns the local class distribution with the global class distribution and restores the recognition ability of tail classes; thus, class-distribution heterogeneity is alleviated. The EFM strategy alleviates object-appearance heterogeneity by using inter and intra-contrastive loss. Additionally, LoGo is a balancing module to preserve local personality and global generalization during the collaboration. Extensive experiments on three datasets show that our GeoFed consistently outperforms the current state-of-the-art methods. Overall, the analyses and comparisons provided in this paper show that the proposed GeoFed framework effectively promotes institutions' cooperation. Meanwhile, geographic heterogeneity is a common issue in various FL tasks for remote sensing. This bodes well for GeoFed's broad applicability across other remote sensing interpretation tasks (e.g., Change Detection, Object Detection and Scene Classification).

In the future, we hope our framework will inspire further research on privacy-preserving collaborative learning in remote sensing. Due to diverse computational resources and variations in data volume among institutions, a well-designed scheduling strategy is essential to avoid resource wastage. In addition, our GeoFed framework focuses on geographic heterogeneity from the data perspective. Thus, it is currently tailored for homogeneous model settings. It can be extended to accommodate heterogeneous model scenarios with some delicate design, such as knowledge distillation. Our future work will handle the above challenges and consider the real application of GeoFed to global-scale remote sensing data.

\bibliographystyle{IEEEtran}

\bibliography{ref}

\begin{thebibliography}{10}
\providecommand{\url}[1]{#1}
\csname url@samestyle\endcsname
\providecommand{\newblock}{\relax}
\providecommand{\bibinfo}[2]{#2}
\providecommand{\BIBentrySTDinterwordspacing}{\spaceskip=0pt\relax}
\providecommand{\BIBentryALTinterwordstretchfactor}{4}
\providecommand{\BIBentryALTinterwordspacing}{\spaceskip=\fontdimen2\font plus
\BIBentryALTinterwordstretchfactor\fontdimen3\font minus \fontdimen4\font\relax}
\providecommand{\BIBforeignlanguage}[2]{{%
\expandafter\ifx\csname l@#1\endcsname\relax
\typeout{** WARNING: IEEEtran.bst: No hyphenation pattern has been}%
\typeout{** loaded for the language `#1'. Using the pattern for}%
\typeout{** the default language instead.}%
\else
\language=\csname l@#1\endcsname
\fi
#2}}
\providecommand{\BIBdecl}{\relax}
\BIBdecl

\bibitem{ISPRS_LULC}
Y.~Wang, Y.~Sun, X.~Cao, Y.~Wang, W.~Zhang, and X.~Cheng, ``A review of regional and global scale land use/land cover (lulc) mapping products generated from satellite remote sensing,'' \emph{ISPRS Journal of Photogrammetry and Remote Sensing}, vol. 206, pp. 311--334, 2023.

\bibitem{RSE2024deepULU}
Z.~Li, B.~Chen, S.~Wu, M.~Su, J.~M. Chen, and B.~Xu, ``Deep learning for urban land use category classification: A review and experimental assessment,'' \emph{Remote Sensing of Environment}, vol. 311, p. 114290, 2024.

\bibitem{RSE_global_landuse}
Y.~Zhong, B.~Yan, J.~Yi, R.~Yang, M.~Xu, Y.~Su, Z.~Zheng, and L.~Zhang, ``Global urban high-resolution land-use mapping: From benchmarks to multi-megacity applications,'' \emph{Remote Sensing of Environment}, vol. 298, p. 113758, 2023.

\bibitem{GRSM_Aisecurity}
Y.~Xu, T.~Bai, W.~Yu, S.~Chang, P.~M. Atkinson, and P.~Ghamisi, ``Ai security for geoscience and remote sensing: Challenges and future trends,'' \emph{IEEE Geoscience and Remote Sensing Magazine}, vol.~11, no.~2, pp. 60--85, 2023.

\bibitem{GRSM_FL}
B.~Büyüktas, G.~Sumbul, and B.~Demir, ``Federated learning across decentralized and unshared archives for remote sensing image classification: A review,'' \emph{IEEE Geoscience and Remote Sensing Magazine}, pp. 2--18, 2024.

\bibitem{RSE_global_scale}
Z.~Zheng, J.~Yu, X.~Zhang, and S.~Du, ``Development of a 30 m resolution global sand dune/sheet classification map (gsds30) using multi-source remote sensing data,'' \emph{Remote Sensing of Environment}, vol. 302, p. 113973, 2024.

\bibitem{IJDE_national_scale}
Y.~Wang, Y.~Xu, X.~Xu, X.~Jiang, Y.~Mo, H.~Cui, S.~Zhu, and H.~Wu, ``Evaluation of six global high-resolution global land cover products over china,'' \emph{International Journal of Digital Earth}, vol.~17, no.~1, p. 2301673, 2024.

\bibitem{RSE_region_scale}
A.~Rizayeva, M.~D. Nita, and V.~C. Radeloff, ``Large-area, 1964 land cover classifications of corona spy satellite imagery for the caucasus mountains,'' \emph{Remote Sensing of Environment}, vol. 284, p. 113343, 2023.

\bibitem{ISPRS_trustworthy}
S.~Wang, W.~Han, X.~Huang, X.~Zhang, L.~Wang, and J.~Li, ``Trustworthy remote sensing interpretation: Concepts, technologies, and applications,'' \emph{ISPRS Journal of Photogrammetry and Remote Sensing}, vol. 209, pp. 150--172, 2024.

\bibitem{AISTATS_FedAvg}
B.~McMahan, E.~Moore, D.~Ramage, S.~Hampson, and B.~A. y~Arcas, ``Communication-efficient learning of deep networks from decentralized data,'' in \emph{Artificial intelligence and statistics}.\hskip 1em plus 0.5em minus 0.4em\relax PMLR, 2017, pp. 1273--1282.

\bibitem{shao2024selective}
J.~Shao, F.~Wu, and J.~Zhang, ``Selective knowledge sharing for privacy-preserving federated distillation without a good teacher,'' \emph{Nature Communications}, vol.~15, no.~1, p. 349, 2024.

\bibitem{second_law}
M.~F. Goodchild, ``The validity and usefulness of laws in geographic information science and geography,'' \emph{Annals of the Association of American Geographers}, vol.~94, no.~2, pp. 300--303, 2004.

\bibitem{ISPRS_landuse_hetero}
D.~Schulz, H.~Yin, B.~Tischbein, S.~Verleysdonk, R.~Adamou, and N.~Kumar, ``Land use mapping using sentinel-1 and sentinel-2 time series in a heterogeneous landscape in niger, sahel,'' \emph{ISPRS Journal of Photogrammetry and Remote Sensing}, vol. 178, pp. 97--111, 2021.

\bibitem{RSE_hetero_landscape}
Z.~Zheng, S.~Du, Y.-C. Wang, and Q.~Wang, ``Mining the regularity of landscape-structure heterogeneity to improve urban land-cover mapping,'' \emph{Remote Sensing of Environment}, vol. 214, pp. 14--32, 2018.

\bibitem{CVPR_re_hetero}
M.~Mendieta, T.~Yang, P.~Wang, M.~Lee, Z.~Ding, and C.~Chen, ``Local learning matters: Rethinking data heterogeneity in federated learning,'' in \emph{Proceedings of the IEEE/CVF Conference on Computer Vision and Pattern Recognition (CVPR)}, June 2022, pp. 8397--8406.

\bibitem{MLSys_FedProx}
T.~Li, A.~K. Sahu, M.~Zaheer, M.~Sanjabi, A.~Talwalkar, and V.~Smith, ``Federated optimization in heterogeneous networks,'' in \emph{Proceedings of Machine Learning and Systems}, I.~Dhillon, D.~Papailiopoulos, and V.~Sze, Eds., vol.~2, 2020, pp. 429--450.

\bibitem{CVPR_FedSeg}
J.~Miao, Z.~Yang, L.~Fan, and Y.~Yang, ``Fedseg: Class-heterogeneous federated learning for semantic segmentation,'' in \emph{Proceedings of the IEEE/CVF Conference on Computer Vision and Pattern Recognition (CVPR)}, June 2023, pp. 8042--8052.

\bibitem{ECCV_flat}
D.~Caldarola, B.~Caputo, and M.~Ciccone, ``Improving generalization in federated learning by seeking flat minima,'' in \emph{European Conference on Computer Vision}.\hskip 1em plus 0.5em minus 0.4em\relax Springer, 2022, pp. 654--672.

\bibitem{TPAMI_LT_survey}
Y.~Zhang, B.~Kang, B.~Hooi, S.~Yan, and J.~Feng, ``Deep long-tailed learning: A survey,'' \emph{IEEE Transactions on Pattern Analysis and Machine Intelligence}, vol.~45, no.~9, pp. 10\,795--10\,816, 2023.

\bibitem{ICCV_ls_semi}
R.~Dong, L.~Mou, M.~Chen, W.~Li, X.-Y. Tong, S.~Yuan, L.~Zhang, J.~Zheng, X.~Zhu, and H.~Fu, ``Large-scale land cover mapping with fine-grained classes via class-aware semi-supervised semantic segmentation,'' in \emph{Proceedings of the IEEE/CVF International Conference on Computer Vision (ICCV)}, October 2023, pp. 16\,783--16\,793.

\bibitem{ESSD_SinoLC1}
Z.~Li, W.~He, M.~Cheng, J.~Hu, G.~Yang, and H.~Zhang, ``Sinolc-1: the first 1 m resolution national-scale land-cover map of china created with a deep learning framework and open-access data,'' \emph{Earth System Science Data}, vol.~15, no.~11, pp. 4749--4780, 2023.

\bibitem{ISPRS_FBP}
X.-Y. Tong, G.-S. Xia, and X.~X. Zhu, ``Enabling country-scale land cover mapping with meter-resolution satellite imagery,'' \emph{ISPRS Journal of Photogrammetry and Remote Sensing}, vol. 196, pp. 178--196, 2023.

\bibitem{RSE_LCZ_mapping}
F.~Huang, S.~Jiang, W.~Zhan, B.~Bechtel, Z.~Liu, M.~Demuzere, Y.~Huang, Y.~Xu, L.~Ma, W.~Xia \emph{et~al.}, ``Mapping local climate zones for cities: A large review,'' \emph{Remote Sensing of Environment}, vol. 292, p. 113573, 2023.

\bibitem{CVPR_dynamic}
A.~Toker, L.~Kondmann, M.~Weber, M.~Eisenberger, A.~Camero, J.~Hu, A.~P. Hoderlein, c.~\c{S}enaras, T.~Davis, D.~Cremers, G.~Marchisio, X.~X. Zhu, and L.~Leal-Taix\'e, ``Dynamicearthnet: Daily multi-spectral satellite dataset for semantic change segmentation,'' in \emph{Proceedings of the IEEE/CVF Conference on Computer Vision and Pattern Recognition (CVPR)}, June 2022, pp. 21\,158--21\,167.

\bibitem{ACM_heter}
M.~Ye, X.~Fang, B.~Du, P.~C. Yuen, and D.~Tao, ``Heterogeneous federated learning: State-of-the-art and research challenges,'' \emph{ACM Computing Surveys}, vol.~56, no.~3, pp. 1--44, 2023.

\bibitem{ICLR_convergence}
X.~Li, K.~Huang, W.~Yang, S.~Wang, and Z.~Zhang, ``On the convergence of fedavg on non-iid data,'' \emph{arXiv preprint arXiv:1907.02189}, 2019.

\bibitem{IJCAI_CReFF}
X.~Shang, Y.~Lu, G.~Huang, and H.~Wang, ``Federated learning on heterogeneous and long-tailed data via classifier re-training with federated features,'' in \emph{Proceedings of the Thirty-First International Joint Conference on Artificial Intelligence, {IJCAI-22}}, L.~D. Raedt, Ed., 7 2022, pp. 2218--2224.

\bibitem{ICDE_experi}
Q.~Li, Y.~Diao, Q.~Chen, and B.~He, ``Federated learning on non-iid data silos: An experimental study,'' in \emph{2022 IEEE 38th International Conference on Data Engineering (ICDE)}, 2022, pp. 965--978.

\bibitem{yuan2024communication}
L.~Yuan, D.-J. Han, S.~Wang, D.~Upadhyay, and C.~G. Brinton, ``Communication-efficient multimodal federated learning: Joint modality and client selection,'' \emph{arXiv preprint arXiv:2401.16685}, 2024.

\bibitem{CVPR_FISS}
J.~Dong, D.~Zhang, Y.~Cong, W.~Cong, H.~Ding, and D.~Dai, ``Federated incremental semantic segmentation,'' in \emph{Proceedings of the IEEE/CVF Conference on Computer Vision and Pattern Recognition (CVPR)}, June 2023, pp. 3934--3943.

\bibitem{CVPR_FedDG}
Q.~Liu, C.~Chen, J.~Qin, Q.~Dou, and P.-A. Heng, ``Feddg: Federated domain generalization on medical image segmentation via episodic learning in continuous frequency space,'' in \emph{Proceedings of the IEEE/CVF Conference on Computer Vision and Pattern Recognition (CVPR)}, June 2021, pp. 1013--1023.

\bibitem{CVPR_DaFKD}
H.~Wang, Y.~Li, W.~Xu, R.~Li, Y.~Zhan, and Z.~Zeng, ``Dafkd: Domain-aware federated knowledge distillation,'' in \emph{Proceedings of the IEEE/CVF Conference on Computer Vision and Pattern Recognition (CVPR)}, June 2023, pp. 20\,412--20\,421.

\bibitem{CVPR_Style_DG}
W.~Huang, C.~Chen, Y.~Li, J.~Li, C.~Li, F.~Song, Y.~Yan, and Z.~Xiong, ``Style projected clustering for domain generalized semantic segmentation,'' in \emph{Proceedings of the IEEE/CVF Conference on Computer Vision and Pattern Recognition (CVPR)}, June 2023, pp. 3061--3071.

\bibitem{IROS_auto_drive}
L.~Fantauzzo, E.~Fanì, D.~Caldarola, A.~Tavera, F.~Cermelli, M.~Ciccone, and B.~Caputo, ``Feddrive: Generalizing federated learning to semantic segmentation in autonomous driving,'' in \emph{2022 IEEE/RSJ International Conference on Intelligent Robots and Systems (IROS)}, 2022, pp. 11\,504--11\,511.

\bibitem{ICCV_GBME_FedLT}
Y.~Zeng, L.~Liu, L.~Liu, L.~Shen, S.~Liu, and B.~Wu, ``Global balanced experts for federated long-tailed learning,'' in \emph{Proceedings of the IEEE/CVF International Conference on Computer Vision (ICCV)}, October 2023, pp. 4815--4825.

\bibitem{IOT_FedMargin}
U.~Michieli, M.~Toldo, and M.~Ozay, ``Federated learning via attentive margin of semantic feature representations,'' \emph{IEEE Internet of Things Journal}, vol.~10, no.~2, pp. 1517--1535, 2023.

\bibitem{CVPR_rethinking_prototype}
W.~Huang, M.~Ye, Z.~Shi, H.~Li, and B.~Du, ``Rethinking federated learning with domain shift: A prototype view,'' in \emph{Proceedings of the IEEE/CVF Conference on Computer Vision and Pattern Recognition (CVPR)}, June 2023, pp. 16\,312--16\,322.

\bibitem{CVPR_re_archi}
L.~Qu, Y.~Zhou, P.~P. Liang, Y.~Xia, F.~Wang, E.~Adeli, L.~Fei-Fei, and D.~Rubin, ``Rethinking architecture design for tackling data heterogeneity in federated learning,'' in \emph{Proceedings of the IEEE/CVF Conference on Computer Vision and Pattern Recognition (CVPR)}, June 2022, pp. 10\,061--10\,071.

\bibitem{ICCV_FRAug}
H.~Chen, A.~Frikha, D.~Krompass, J.~Gu, and V.~Tresp, ``Fraug: Tackling federated learning with non-iid features via representation augmentation,'' in \emph{Proceedings of the IEEE/CVF International Conference on Computer Vision (ICCV)}, October 2023, pp. 4849--4859.

\bibitem{CVPR_Elastic}
D.~Chen, J.~Hu, V.~J. Tan, X.~Wei, and E.~Wu, ``Elastic aggregation for federated optimization,'' in \emph{Proceedings of the IEEE/CVF Conference on Computer Vision and Pattern Recognition (CVPR)}, June 2023, pp. 12\,187--12\,197.

\bibitem{Zhang_2024_CVPR}
J.~Zhang, Y.~Liu, Y.~Hua, and J.~Cao, ``An upload-efficient scheme for transferring knowledge from a server-side pre-trained generator to clients in heterogeneous federated learning,'' in \emph{Proceedings of the IEEE/CVF Conference on Computer Vision and Pattern Recognition (CVPR)}, June 2024, pp. 12\,109--12\,119.

\bibitem{chen2024free}
S.~Chen, T.~Shu, H.~Zhao, J.~Wang, S.~Ren, and L.~Yang, ``Free lunch for federated remote sensing target fine-grained classification: A parameter-efficient framework,'' \emph{arXiv preprint arXiv:2401.01493}, 2024.

\bibitem{ACM_aerial}
P.~Chhikara, R.~Tekchandani, N.~Kumar, and S.~Tanwar, ``Federated learning-based aerial image segmentation for collision-free movement and landing,'' in \emph{Proceedings of the 4th ACM MobiCom Workshop on Drone Assisted Wireless Communications for 5G and Beyond}, 2021, pp. 13--18.

\bibitem{IoT_FL_sky}
Y.~Liu, J.~Nie, X.~Li, S.~H. Ahmed, W.~Y.~B. Lim, and C.~Miao, ``Federated learning in the sky: Aerial-ground air quality sensing framework with uav swarms,'' \emph{IEEE Internet of Things Journal}, vol.~8, no.~12, pp. 9827--9837, 2021.

\bibitem{RS_DP}
Z.~Zhang, X.~Ma, and J.~Ma, ``Local differential privacy based membership-privacy-preserving federated learning for deep-learning-driven remote sensing,'' \emph{Remote Sensing}, vol.~15, no.~20, p. 5050, 2023.

\bibitem{TGRS_fedhyper}
W.~Cai, M.~Gao, Y.~Ding, X.~Ning, X.~Bai, and P.~Qian, ``Stereo attention cross-decoupling fusion-guided federated neural learning for hyperspectral image classification,'' \emph{IEEE Transactions on Geoscience and Remote Sensing}, vol.~61, pp. 1--16, 2023.

\bibitem{JSTARS_FL_image_classification}
P.~Tam, S.~Math, C.~Nam, and S.~Kim, ``Adaptive resource optimized edge federated learning in real-time image sensing classifications,'' \emph{IEEE Journal of Selected Topics in Applied Earth Observations and Remote Sensing}, vol.~14, pp. 10\,929--10\,940, 2021.

\bibitem{TGRS_Radar}
Y.~Song and G.~Dong, ``Federated target recognition for multiradar sensor data security,'' \emph{IEEE Transactions on Geoscience and Remote Sensing}, vol.~61, pp. 1--10, 2023.

\bibitem{TGRS_FedPM}
X.~Zhang, B.~Zhang, W.~Yu, and X.~Kang, ``Federated deep learning with prototype matching for object extraction from very-high-resolution remote sensing images,'' \emph{IEEE Transactions on Geoscience and Remote Sensing}, vol.~61, pp. 1--16, 2023.

\bibitem{IGARSS_proto_IAIL}
B.~Zhang, X.~Zhang, M.-O. Pun, and M.~Liu, ``Prototype-based clustered federated learning for semantic segmentation of aerial images,'' in \emph{IGARSS 2022 - 2022 IEEE International Geoscience and Remote Sensing Symposium}, 2022, pp. 2227--2230.

\bibitem{ICO_FSS}
Y.~Khasgiwala, D.~T. Castellino, and S.~Deshmukh, ``A decentralized federated learning paradigm for semantic segmentation of geospatial data,'' in \emph{Intelligent Computing \& Optimization: Proceedings of the 4th International Conference on Intelligent Computing and Optimization 2021 (ICO2021) 3}.\hskip 1em plus 0.5em minus 0.4em\relax Springer, 2022, pp. 196--206.

\bibitem{GRSL_FS}
L.~Hu, P.~Cheng, Y.~Wang, Z.~Wang, K.~Chen, X.~Sun, and D.~Zhang, ``Fs-dcl: Distributed collaborative learning for few-shot remote sensing image classification,'' \emph{IEEE Geoscience and Remote Sensing Letters}, 2023.

\bibitem{TGRS_MPL}
J.~Li, M.~Gong, Z.~Liu, S.~Wang, Y.~Zhang, Y.~Zhou, and Y.~Gao, ``Towards multi-party personalized collaborative learning in remote sensing,'' \emph{IEEE Transactions on Geoscience and Remote Sensing}, 2024.

\bibitem{Arxiv_FedDiff}
D.~Li, W.~Xie, Z.~Wang, Y.~Lu, Y.~Li, and L.~Fang, ``Feddiff: Diffusion model driven federated learning for multi-modal and multi-clients,'' \emph{arXiv preprint arXiv:2401.02433}, 2023.

\bibitem{ACM_MPC}
D.~Byrd and A.~Polychroniadou, ``Differentially private secure multi-party computation for federated learning in financial applications,'' in \emph{Proceedings of the First ACM International Conference on AI in Finance}, 2020, pp. 1--9.

\bibitem{CVPR_EWF}
J.-W. Xiao, C.-B. Zhang, J.~Feng, X.~Liu, J.~van~de Weijer, and M.-M. Cheng, ``Endpoints weight fusion for class incremental semantic segmentation,'' in \emph{Proceedings of the IEEE/CVF Conference on Computer Vision and Pattern Recognition (CVPR)}, June 2023, pp. 7204--7213.

\bibitem{ICCV_MAP}
M.~Siam, B.~N. Oreshkin, and M.~Jagersand, ``Amp: Adaptive masked proxies for few-shot segmentation,'' in \emph{Proceedings of the IEEE/CVF International Conference on Computer Vision (ICCV)}, October 2019, p.~1.

\bibitem{ICML_InfoNCE}
T.~Chen, S.~Kornblith, M.~Norouzi, and G.~Hinton, ``A simple framework for contrastive learning of visual representations,'' in \emph{International conference on machine learning}.\hskip 1em plus 0.5em minus 0.4em\relax PMLR, 2020, pp. 1597--1607.

\bibitem{ISPRS_CASID}
S.~Liu, L.~Chen, L.~Zhang, J.~Hu, and Y.~Fu, ``A large-scale climate-aware satellite image dataset for domain adaptive land-cover semantic segmentation,'' \emph{ISPRS Journal of Photogrammetry and Remote Sensing}, vol. 205, pp. 98--114, 2023.

\bibitem{IGARSS_IAIL}
E.~Maggiori, Y.~Tarabalka, G.~Charpiat, and P.~Alliez, ``Can semantic labeling methods generalize to any city? the inria aerial image labeling benchmark,'' in \emph{2017 IEEE International Geoscience and Remote Sensing Symposium (IGARSS)}, 2017, pp. 3226--3229.

\bibitem{ECCV_DeepLabV3p}
L.-C. Chen, Y.~Zhu, G.~Papandreou, F.~Schroff, and H.~Adam, ``Encoder-decoder with atrous separable convolution for semantic image segmentation,'' in \emph{Computer Vision -- ECCV 2018}, V.~Ferrari, M.~Hebert, C.~Sminchisescu, and Y.~Weiss, Eds.\hskip 1em plus 0.5em minus 0.4em\relax Cham: Springer International Publishing, 2018, pp. 833--851.

\bibitem{CVPR_ResNet}
K.~He, X.~Zhang, S.~Ren, and J.~Sun, ``Deep residual learning for image recognition,'' in \emph{2016 IEEE Conference on Computer Vision and Pattern Recognition (CVPR)}, 2016, pp. 770--778.

\bibitem{CVPR_ImageNet}
J.~Deng, W.~Dong, R.~Socher, L.-J. Li, K.~Li, and L.~Fei-Fei, ``Imagenet: A large-scale hierarchical image database,'' in \emph{2009 IEEE Conference on Computer Vision and Pattern Recognition}, 2009, pp. 248--255.

\bibitem{CVPR_Moon}
Q.~Li, B.~He, and D.~Song, ``Model-contrastive federated learning,'' in \emph{Proceedings of the IEEE/CVF Conference on Computer Vision and Pattern Recognition (CVPR)}, June 2021, pp. 10\,713--10\,722.

\bibitem{ICML_FedDisco}
R.~Ye, M.~Xu, J.~Wang, C.~Xu, S.~Chen, and Y.~Wang, ``Feddisco: federated learning with discrepancy-aware collaboration,'' in \emph{Proceedings of the 40th International Conference on Machine Learning}, ser. ICML'23.\hskip 1em plus 0.5em minus 0.4em\relax JMLR.org, 2023, p.~24.

\bibitem{ICCV_GradCAM}
R.~R. Selvaraju, M.~Cogswell, A.~Das, R.~Vedantam, D.~Parikh, and D.~Batra, ``Grad-cam: Visual explanations from deep networks via gradient-based localization,'' in \emph{2017 IEEE International Conference on Computer Vision (ICCV)}, 2017, pp. 618--626.

\end{thebibliography}

\end{document}